%% file: rohrbach16chapter.tex
\pdfoutput=1 
\documentclass[graybox]{svmult}

\usepackage{mathptmx}       %
\usepackage{helvet}         %
\usepackage{courier}        %
\usepackage{type1cm}        %
\usepackage{makeidx}         %
\usepackage{graphicx}        %
\usepackage{multicol}        %
\usepackage[bottom]{footmisc}%
\usepackage{xspace}
\usepackage{array}
\usepackage{booktabs}
\usepackage[english]{babel}
\include{macros}

\include{defines}

\renewcommand{\tabcolsep}{6pt}
\usepackage[square,numbers]{natbib}
\makeindex             %

\begin{document}

\title*{Attributes for Interaction between Natural Language and Computer Vision}
\title*{Attributes as Semantic Units between \newline Natural Language and Visual Recognition}
\titlerunning{Attributes as Semantic Units between Natural Language and Visual Recognition} %
\author{Marcus Rohrbach}
\institute{Marcus Rohrbach \at UC Berkley EECS and ICSI, Berkeley, USA}
\maketitle

\newcommand{\myabstract}{Impressive progress has been made in the fields of computer vision and natural language processing. However, it remains a challenge to find the best point of interaction for these very different modalities. In this chapter we discuss how attributes allow us to exchange information between the two modalities and in this way lead to an interaction on a semantic level. Specifically we discuss how attributes allow using knowledge mined from language resources for recognizing novel visual categories, how we can generate sentence description about images and video, how we can ground natural language in visual content, and finally, how we can answer natural language questions about images.
}

\abstract*{\myabstract}

\abstract{\myabstract}

\input{introduction}

\input{novelobjects}

\input{description}
\input{novelimgdesc}
\input{grounding}

\input{qa}

\section{Conclusions}
\label{sec:conclusions}
In this chapter we presented several tasks and approaches where attributes enable a connection of visual recognition with natural language on a semantic level. For recognizing novel object categories or activities, attribute can build an intermediate representation which allows incorporating knowledge mined from language resources or script data (\Secref{sec:objectrecognition}). For this scenario we saw that semantic attribute classifiers additionally build a good metric distance space useful for constructing instance graphs and learning composite activity recognition models. In \Secref{sec:description} we explained how an intermediate level of attributes can be used to describe videos  with multiple sentences and at a variable level and allow describing  novel object categories.  In \Secref{sec:ground} we presented approaches for unsupervised and supervised grounding of phrases in images. Different phrases are semantically overlapping and the examined approaches try to relate these semantic units by jointly learning representations for the visual and language modalities.  \Secref{sec:qa}  discusses an approach to visual question answering which composes the most important attributes of a question in a compositional computation graph, whose parameters are learned end-to-end only by back-propagating from the answers.  

While the discussed approaches take a step towards the challenges discussed in \Secref{sec:intro:challenges}, there are many future steps ahead. While the approaches in \Secref{sec:objectrecognition} use many advanced semantic relatedness measures minded from diverse language resources they are not jointly trained on textual and visual modalities. \citet{regneri13tacl} and \citet{silberer13acl}
take a step in this direction by looking at joint semantic representation from the textual and visual modalities.  \Secref{sec:description} presents compositional models for describing videos, but it is only a first step towards automatically describing a movie to a blind person as humans can do it \cite{rohrbach15cvpr}, which will require an even higher degree of semantic understanding, and transfer within and between modalities. \Secref{sec:ground} describes interesting ideas to grounding in images and it will be interesting to see how this scales to the size of the Internet. Visual question answering (\Secref{sec:qa}) is an interesting emerging direction with many challenges  as it  requires to solve all of the above, at least to some extend.

\begin{acknowledgement}
I would like to thank all my co-authors, especially those whose publications are presented in this chapter. Namely, Sikandar Amin, Jacob Andreas, %
Mykhaylo Andriluka, 
Trevor Darrell, %
Sandra Ebert,  
Jiashi Feng, %
Annemarie Friedrich, 
Iryna Gurevych, 
Lisa Anne Hendricks, %
Ronghang Hu, %
Dan Klein, %
Raymond Mooney, %
Manfred Pinkal,  
Wei Qiu, 
Michaela Regneri,
Anna Rohrbach, %
Kate Saenko, %
Michael Stark, %
Bernt Schiele, %
Gy{\"o}rgy Szarvas,
Stefan Thater,  
Ivan Titov, 
Subhashini Venugopalan, %
and Huazhe Xu. %

Marcus Rohrbach was supported by a fellowship within the FITweltweit-Program of the German Academic Exchange Service (DAAD).
\end{acknowledgement}

\bibliography{biblioLong,related,rohrbach}

\end{document}

%% file: macros.tex
\newcommand{\approach}{GroundeR\xspace}

\newcommand{\mod}[1]{{\small\texttt{#1}}}
\newcommand{\nmn}{Neural Module Networks\xspace}

\newcommand{\Scriptknowledge}{Script data\xspace}
\newcommand{\tfidf}{{tf$*$idf}\xspace}

\newcommand{\ApproachName}[1]{Propagated Semantic Transfer}

%% file: defines.tex
\graphicspath{{./fig/}{.}{./fig/description/}{./fig/novelimgdesc/}{./fig/grounding/}}
\usepackage{multirow} 
 \makeatletter
 \DeclareRobustCommand\onedot{\futurelet\@let@token\@onedot}
 \def\@onedot{\ifx\@let@token.\else.\null\fi\xspace}
 \def\eg{e.g\onedot} 
 \def\ie{i.e\onedot} 
  
  \def\vs{vs\onedot}

 \def\vs{vs\onedot}
 \makeatother

\DeclareRobustCommand{\figref}[1]{Fig\onedot~\ref{#1}}

\DeclareRobustCommand{\Figref}[1]{Fig\onedot~\ref{#1}}
\DeclareRobustCommand{\Figsref}[1]{Figures~\ref{#1}}

\DeclareRobustCommand{\Secref}[1]{Section~\ref{#1}}
\DeclareRobustCommand{\secref}[1]{Section~\ref{#1}}

\DeclareRobustCommand{\secsref}[1]{Sections~\ref{#1}}

\DeclareRobustCommand{\Tableref}[1]{Table~\ref{#1}}

%% file: introduction.tex
\section{Introduction}
\label{sec:introduction}
Computer vision has made impressive progress in recognizing large number of objects categories \cite{szegedy15cvpr}, diverse activities \cite{wang13iccv}, and most recently also in describing images and videos with natural language sentences \cite{vinyals15cvpr,venugopalan15iccv} and answering natural language questions about images \cite{malinowski14nips}.
Given sufficient training data these approaches can achieve impressive performance, sometimes even on par with humans \cite{he15iccv}. However, humans have two key abilities most computer vision system lack. On the one hand humans can easily generalize to novel categories with no or very little training data. 
On the other hand, humans can rely on other modalities, most notably language, to incorporate knowledge in the recognition process. To do so humans seem to be able to rely on compositionality and transferability, which means they can break up complex problems into components, and use previously learned components in other (recognition) tasks.
In this chapter we discuss how attributes can form such components which allow to transfer and share knowledge, incorporate external linguistic knowledge, and decompose the challenging problems of visual description and question answering into smaller semantic units, which are easier to recognize and associate with textual representation.

\begin{figure}[t]
\begin{center}
\includegraphics[width=0.7\textwidth]{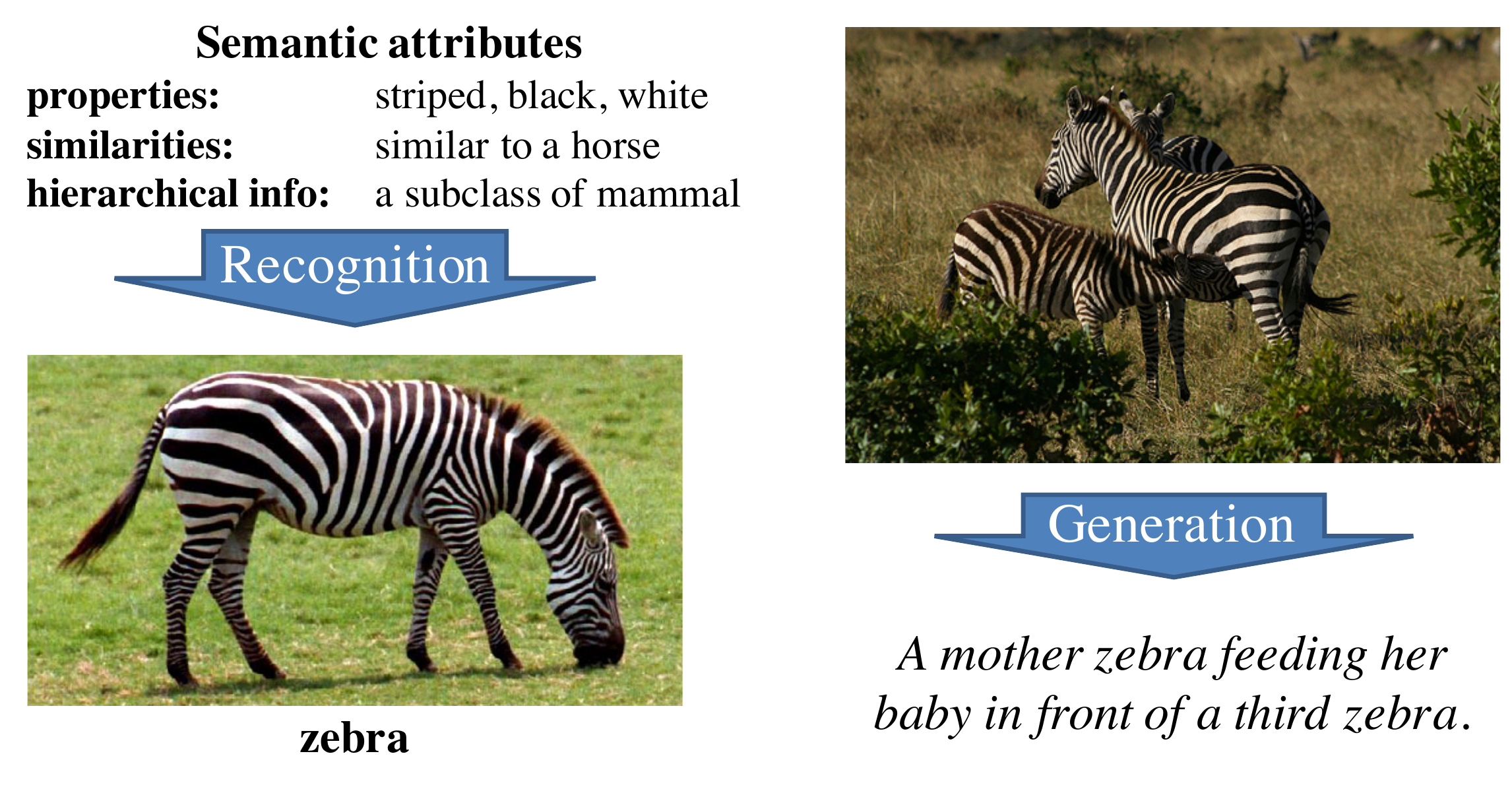}
\begin{tabular}{p{5cm} @{\hspace{0.5cm}} p{5cm} }
(a) Semantic attributes allow recognition of novel classes. &
(b) Sentence description for an image. \newline Image and caption from MS COCO \cite{chen15arXiv1504.00325}.
\end{tabular}
  \caption{Examples for textual descriptions and visual content.}
  \label{fig:intro:teaser}
\end{center}
\end{figure}

Let us first illustrate this with two examples. Attribute descriptions given in the form of hierarchical information (\emph{a mammal}), properties (\emph{striped, black, and white}), and similarities, (\emph{similar to a horse}), allow humans to recognize a visual category, even if they never observed this category before. Given this description in form of attributes most humans would be able to recognize the animal shown in \figref{fig:intro:teaser}(a) as a \emph{zebra}. 
Furthermore, once humans know that \figref{fig:intro:teaser}(a) is a \emph{zebra}, they can describe what it is doing within a natural sentence, even if they never saw example images with captions of zebras before (\figref{fig:intro:teaser}b). A promising way to handle these challenges is to have compositional models which allow interaction between multi-modal information at a semantic level.

One prominent way to model such a semantic level are semantic attributes. As the term ``\emph{attribute}'' has a large variety of definitions in the computer vision literature we define for the course of this chapter as follows. 
\begin{definition}
An \emph{attribute} is a semantic unit, which has a visual and a textual representation. 
\end{definition}
The first part of this definition, the restriction to a semantic unit is important to discriminate attributes from other representations, which do not have human interpretable meaning, such as image gradients, bag of (visual) words, or hidden representations in deep neural networks. We will refer to these as \emph{features}. Of course for a specific feature, one can try to find or associate it with a semantic meaning or unit, but typically it is unknown and once one is able to identify such a association, one has found a representation for this semantic attribute. The restriction to a semantic unit allows to connect to other sources of information on a semantic level, \ie a level of meaning. In the second part of the definition we restrict it to semantic units which can be both represented textually and visually.\footnote{There are attributes / semantic units, which are not visual but textually, \eg smells, tastes, tactile sensory inputs, and ones which are visual but not textual, which are naturally difficult to describe in language, but think of many visual patterns beyond \emph{striped} and \emph{dotted}, for which we do not have name, or the different visual attributes between two people or faces which humans can clearly recognize but which might be difficult to put into words. We also like to note that some datasets such as Animals with Attributes \cite{lampert13pami} include non-visual attributes, \eg \emph{smelly}, which might still improve classification performance as they are correlated to visual features.} This this specific for this chapter as we want to exploit the connection between language and visual recognition.  From this definition it should also be clear that attributes are not distinct from objects, but rather that objects are also attributes, as they obviously are semantic and have a textual and visual representation.  		

In this chapter we discuss some of the most prominent directions where language understanding and visual recognition interact. Namely how knowledge mined from language resources can help visual recognition%
, how we can ground language in visual content%
, how we can generate language about visual content%
, and finally how we can answer natural language questions about images%
, which can be seen as a combination of grounding the question, recognition, and generating an answer.
It is clear that these directions cannot cover all potential interactions between visual recognition and language. Other directions include generating visual content from language descriptions \citep[\eg][]{zitnick13iccv,liang13tmm} or localizing images in text \ie to find where in a text an image is discussed. %
 In the following we first analyze  challenges for combining visual and linguistic modalities; afterwards %
we provide an overview of this chapter which includes a discussion how the different sections relate to each other and to the idea of attributes.

\subsection{Challenges for combining visual and linguistic modalities}
\label{sec:intro:challenges}
One of the fundamental differences between the visual and the linguistic modality is the level of abstraction. %
The basic data unit of the visual modality is a (photographic) image or video which always shows a specific instance of a category, or even more precisely a certain instance for a specific viewpoint, lighting, pose, time etc. For example \Figref{fig:intro:teaser}(a) shows one specific instance of the category \emph{zebra} from a side view, eating grass.
 In contrast to this, the basic semantic unit of the linguistic modality are words (which are strings of characters or phonemes for spoken language, but we will restrict ourselves to written linguistic expressions in this chapter). Although a word might \emph{refer} to a specific instance, the word, \ie the string, always \emph{represents} a category of objects, activities, or attributes, abstracting from a specific instance. %
 Interestingly this difference, instance versus category level representation, is also what defines  one of the core challenges in visual recognition and is also an important topic in computational linguistics. 
In visual recognition we are interested in defining or learning models which abstract over a specific image or video to understand the visual characteristic of a category. 
In computational linguistics, when automatically parsing a text, we frequently face the inverse challenge of trying to identify intra and extra linguistic references (co-reference resolution / grounding\footnote{\emph{co-reference} is when two or more words refer to the same thing or person within text, while \emph{grounding} looks at how words refer to things outside text, \eg images.}) of a word or phrase. These problems arise because words typically represent concepts rather than instances and because anaphors, synonyms, hypernyms, or metaphorical expressions are used to refer to the identical object in the real world.

Understanding that the visual and linguistic modalities have different levels of abstraction is important when trying to combine both modalities. In \Secref{sec:objectrecognition}  we use linguistic knowledge at category rather than instance level for visual knowledge transfer, \ie we use linguistic knowledge at the level where it is most expressive that is at level of its basic representation. 
In \secref{sec:description}, when describing visual input with natural language, we put the point of interaction at a semantic attribute level and leave concrete realization of sentences to a language model rather than inferring it from the visual representation, \ie we recognize the most important components or attributes of a sentence, which are activities, objects, tools, locations, or scenes and then generate a sentence based on these.
In \secref{sec:ground} we look at a model which grounds phrases which refer to a specific instance by jointly learning visual and textual representations. In \secref{sec:qa} we answer questions about images by learning small modules which recognize visual elements which are selected according to the question and linked to the most important components in the questions, \eg questions words/phrases (\emph{How many}), nouns, (\emph{dog}) and qualifiers (\emph{black}. By this composition in modules or attributes, we create an architecture, which allows learning these attributes, which link visual and textual modality, jointly across all questions and images.

\subsection{Overview and outline}
In this chapter we explain how linguistic knowledge can help to recognize novel object categories and composite activities (\secref{sec:objectrecognition}),  how attributes help to describe videos  and images with natural language sentences (\secref{sec:description}),  how to ground phrases in images  (\secref{sec:ground}), and how compositional computation allows for effective question answering about images (\secref{sec:qa}). We conclude with directions for future work in \secref{sec:conclusions}.

All these directions have in common that attributes form a layer or composition which is beneficial for connecting between textual and visual representations.
In \secref{sec:objectrecognition}, for recognizing novel object categories and composite activities, attributes form the layer where the transfer happens. Attributes are shared across known and novel categories, while information mined from different language resources is able to provide the associations between the know categories and attributes at training time to learn attribute classifiers and between the attributes and novel categories at test time to recognize the novel categories.  

When describing images and videos (\secref{sec:description}), we first learn an intermediate layer of attribute classifiers, which are then used to generate natural language descriptions. This intermediate layer allows us to reason across sentences at a semantic level and in this way to build a model which generates consistent multi-sentence description. Furthermore, we discuss how such an attribute classifier layer allows us to describe novel categories where no paired image-caption data is available.

When grounding sentences in images, we argue that it makes sense to do this on a level of phrases are rather full sentences, as phrases form semantic units, or attributes, which can be well localized in images. Thus, in \secref{sec:ground} we discuss how we localize short phrases or referential expressions in images.

In \secref{sec:qa} we discuss the task of visual question answering which connects these previous sections, as one has to ground the question in the image and then predict or generate an answer. Here we show how we can decompose the question into attributes which are in this case small neural network components, which are composed in a computation graph to predict the answer. This allows us to share and train the attributes across questions and images, but build a neural network which is specific for a given question.

The order of the following sections weakly follows the historic development, where we start with work which appeared at the time when attributes  started to become popular in computer vision \cite{lampert09cvpr,farhadi10cvpr}. And the last section on visual question answering, a problem which requires more complex interactions between language and visual recognition, has only recently become a topic in the computer vision community \cite{malinowski14nips,antol15iccv}.

%% file: novelobjects.tex
\section{Linguistic knowledge for recognition of novel categories}
\label{sec:objectrecognition}

While supervised training is an integral part of building visual, textual, or multi-modal category models, more recently, knowledge transfer between categories has been recognized as an important ingredient to scale to a large number of categories as well as to enable fine-grained categorization. This development reflects the psychological point of view that humans are able to generalize to novel\footnote{We use ``novel'' throughout this chapter to denote categories with no or few labeled training instances.}
categories with only a few training samples \citep{Moses1996,bart05bmvc}.
This has recently gained increased interest in the computer vision and machine learning literature, which look at zero-shot recognition (with no training instances for a class)  \citep{lampert13pami,Farhadi2009,
palatucci09nips,Parikh2011,fu13pami,mensink12eccv,frome13nips}, and one- or few-shot recognition \citep{thrun96nips,bart05bmvc,raina07icml}.
Knowledge transfer is particularly beneficial when scaling to large numbers of classes where training data is limited \citep{mensink12eccv,frome13nips,rohrbach11cvpr}, distinguishing fine-grained categories \citep{Farrell2011,duan12cvpr}, or analyzing compositional activities in videos \citep{fu13pami,rohrbach12eccv}.

 \begin{figure}[t]
 \sidecaption[t]
  \includegraphics[width=0.4\linewidth]{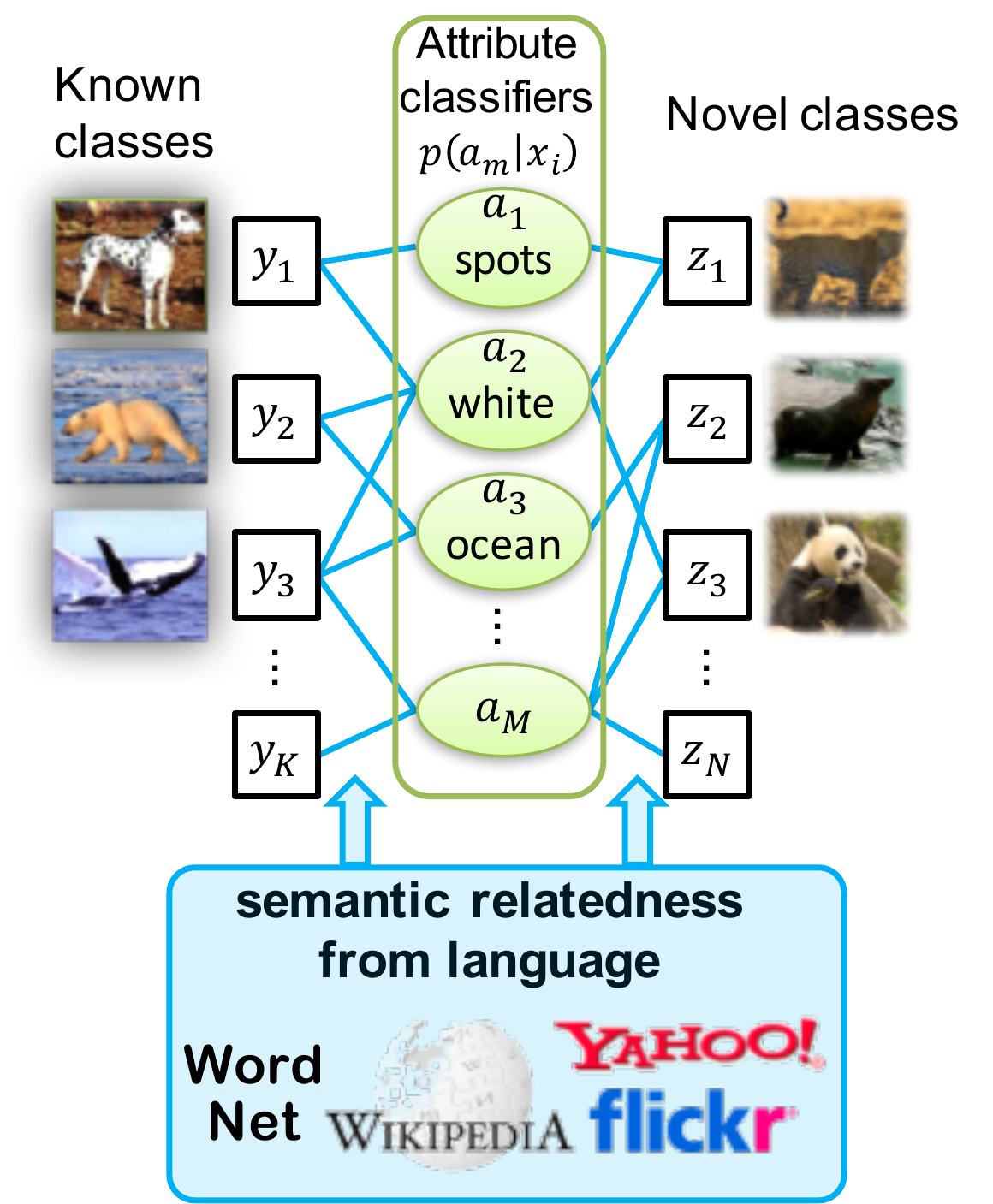}
  \caption{Zero-shot recognition with the Direct Attribute Prediction model \cite{lampert09cvpr} allows recognizing unseen classes $z$ using an intermediate layer of attributes $a$. Instead of manually defined associations between classes and attributes (cyan lines), \citet{rohrbach10cvpr} reduce supervision by mining object-attribute association from language resources, such as Wikipedia, WordNet, and image or web search.}
  \label{fig:attributes}
\end{figure}

Recognizing categories with no or only few labeled training instances is challenging.
In this section we first discuss how we can build attribute classifiers using only category-labeled image data and different language resources which allow recognize novel categories (\secref{sec:attributes}).
And then,  to further improve this transfer learning approach, we discuss how to additionally integrate instance similarity and labeled instances of the novel classes if available (\secref{sec:pst}). Furthermore we discuss what changes have to be made to apply similar ideas to composite activity recognition (\secref{sec:activityrecognition}).

\subsection{Semantic relatedness mined from language resources for zero-shot recognition}
 \label{sec:attributes}

\citet{lampert09cvpr,lampert13pami} propose to use attribute based recognition to allow recognizing unseen categories based on their object-attribute associations. Their Direct Attribute Prediction (DAP) model is visualized in \Figref{fig:attributes}. Given images which are labeled with known category labels $y$ and object-attribute associations $a^y_m$ between categories and attributes, we can learn attribute classifier $p(a_m|x_i)$ for an image $x_i$. This allows to recognize novel categories $z$ if we have associations $a^z_m$.

To scale the approach to a larger number of classes and attributes, \citet{rohrbach10cvpr,rohrbach12lncs,rohrbach11cvpr} show how these previously manual defined attribute associations $a^y_m$ and $a^z_m$ can be replaced with associations mined automatically from different language resources. 
 \Tableref{tbl:novelobjects:zero-shot:results}(a) compares several language resources and measures to estimate semantic relatedness to determine if a class should be associated with a specific attribute. Yahoo Snippets \cite{chen06acl,rohrbach12lncs}, which computes co-occurrence statistics on summary snippets returned by search engines, shows the best performance of all single measures. \citet{rohrbach12lncs} also discuss several fusion strategies to get more robust measures by expanding the attribute inventory with clustering and combining several measures, which can achieve performance on par with manually defined associations (second last versus last line in  \Tableref{tbl:novelobjects:zero-shot:results}a).

\begin{table}[t]
\hspace{-1cm}
\begin{tabular}{p{0.69\textwidth}p{0.40\textwidth}}
\begin{tabular}{llcc}
\toprule
 Language Resource & Measure &  in &   AUC \\
 \cmidrule{1-3}  \cmidrule(rl){4-4}
WordNet \citep{fellbaum:wordnet}, path &Lin measure \cite{lin98icml} &\cite{rohrbach10cvpr}	&60.5\\
 Yahoo Web, hit count	\cite{mihalcea99cl} & Dice coef. \cite{dice45ecology,sorensen48biol}&\cite{rohrbach10cvpr}&60.4\\
Flickr Img, hit count	\cite{rohrbach10cvpr}& Dice coef. \cite{dice45ecology,sorensen48biol}&\cite{rohrbach10cvpr}&70.1\\
Yahoo Img, hit count	\cite{rohrbach10cvpr}& Dice coef. \cite{dice45ecology,sorensen48biol}&\cite{rohrbach10cvpr}&71.0\\
Wikipedia	\cite{rohrbach10cvpr}&ESA  \citep{gabrilovich07ijcai,zesch09jnle}&\cite{rohrbach10cvpr}&69.7\\
Yahoo Snippets	\cite{chen06acl}&Dice/Snippets \cite{rohrbach12lncs}& \cite{rohrbach12lncs}&76.0\\
 \cmidrule(lr){1-1}  \cmidrule(lr){2-2}
Yahoo Img & Expanded attr.  &\cite{rohrbach12lncs}&77.2\\
Combination & Classifier fusion	  &\cite{rohrbach12lncs}&75.9\\
Combination & Expanded attr. 	  &\cite{rohrbach12lncs}&79.5\\
 \cmidrule(lr){1-1}  \cmidrule(lr){2-2}
manual \cite{lampert09cvpr} & &  \cite{rohrbach12lncs} &	79.2\\\bottomrule
\end{tabular}&
\begin{tabular}{l c c@{\ (}c@{)}}
\toprule
 &\multicolumn{3}{c}{ AUC }\\
	\multicolumn{1}{r}{images:}	&  test & \multicolumn{2}{c}{+ train cls$^*$}\\  
\cmidrule(r){1-1}  \cmidrule(lr){2-2}\cmidrule(l){3-4}
\multicolumn{4}{l}{\textbf{Object - Attribute Associations}} \\
Yahoo Img &71.0	& 73.2 & +2.2\\
Classifier fusion& {79.5} & 78.9 & -0.6\\
manual 	& 79.2	& 79.4 & +0.2\\
\cmidrule(r){1-1}  \cmidrule(lr){2-2}\cmidrule(l){3-4}
\multicolumn{4}{l}{\textbf{Direct Similarity}}\\
Yahoo Img 			& {79.9}& {76.4} & -2.5\\
Classifier fusion& 75.9& 72.3 &-3.6\\
\bottomrule
\multicolumn{4}{p{0.38\textwidth}}{$^*$ Effect of adding images from known classes in the test set as distractors/negatives.\vspace{-0.06cm}}
\end{tabular}\\
\vspace{0.05cm}(a) Attribute-based zero-shot recognition. & \vspace{0.05cm}(b) Attributes versus direct-similarity, reported in \cite{rohrbach12lncs}. 
\end{tabular}
\caption{Zero-shot recognition on AwA dataset \cite{lampert09cvpr}. Results for different language resources to mine association. Trained on 92 images per class, mean area under the  ROC curve (AUC) in \%.}
\label{tbl:novelobjects:zero-shot:results}
\end{table}

As an alternative to attributes, \citet{rohrbach10cvpr} also propose to directly transfer information from most similar classes which does not require and intermediate level of attributes.
While this achieves higher performance when the test set only contains novel objects, in the more adversarial settings, when  the test set also contains images from the known categories, the direct similarity based approach significantly drops in performance as can be seen in \Tableref{tbl:novelobjects:zero-shot:results}(b).
 
\begin{table}[t]
\begin{center}
\small
\begin{tabular}{l c c}%
\toprule 
 Approach/Language resource  & in & \multicolumn{1}{c}{Top-5 Error} \\% & \multicolumn{2}{c}{Top-1 Error} \\
\cmidrule(lr){1-1}\cmidrule(lr){2-2} \cmidrule(lr){3-3}
\multicolumn{2}{l}{\textbf{Hierarchy}} \\
leaf WordNet nodes &\cite{rohrbach10cvpr}& 72.8\\% &(4.72)&	91.3 &(11.73)	\\%75.0 & (4.90) &	91.8 &  (11.73)\\
inner WordNet nodes & \cite{rohrbach10cvpr}& 66.7\\% & (4.20) &	88.7 & (11.16)\\%68.7  & (4.36) &	89.7  & (11.29) \\
all WordNet nodes & \cite{rohrbach10cvpr}& 65.2\\% & (4.10)	& 88.4 & (11.24)\\%\textbf{67.3}  & (4.26) & \textbf{89.1}  & (11.26) \\
+ metric learning & \cite{mensink12eccv} & \ \ 64.3$^*$ \\ %
\hline
\multicolumn{2}{l}{\textbf{Part Attributes}} \\
Wikipedia & \cite{rohrbach10cvpr}&80.9\\% & (5.17)& 94.5 & (11.69) \\%87.3 &  (6.20) & 99.5  & (10.77) \\
Yahoo Holonyms & \cite{rohrbach10cvpr}& 77.3 \\%& (4.91) &	94.0 &(12.56)\\%79.9  & (5.09) & 95.1  & (12.70)\\
Yahoo Image & \cite{rohrbach10cvpr}& 81.4 \\%& (5.19)	& 95.5 & (12.53)	\\%94.8  & (5.93) & 99.2  & (14.05)\\
Yahoo Snippets & \cite{rohrbach10cvpr}&76.2\\% & (4.87)	& 93.3 & (11.53) \\
all attributes & \cite{rohrbach10cvpr}& 70.3 \\%& (4.57)	& 90.4 & (11.62)\\
\hline
\multicolumn{2}{l}{\textbf{Direct Similarity}} \\ 
Wikipedia  & \cite{rohrbach10cvpr}& 75.6 \\%& (5.20) & 	91.8 & (11.28)\\%77.0  & (5.27) & 92.0  & (11.38)  \\
Yahoo Web & \cite{rohrbach10cvpr}& 69.3 \\%& (4.49) & 	89.7 & (11.10)\\%70.9  & (4.64) & 90.1  & (11.15) \\
Yahoo Image & \cite{rohrbach10cvpr}& 72.0 \\%& (4.60) & 90.7 & (11.26)\\%74.0  & (4.77) & 91.4  & (11.39)\\
Yahoo Snippets & \cite{rohrbach10cvpr}& 75.5 \\%& (4.89) & 91.6 & (11.27)\\
all measures & \cite{rohrbach10cvpr}& 66.6 \\%& (4.41) & 88.4 & (10.65)\\
 \midrule
 \multicolumn{2}{l}{\textbf{Label embedding}} \\ 
DeViSe & \cite{frome13nips} &	\ \ 68.2$^*$ \\ %
\bottomrule
\end{tabular}
\caption[Large scale zero-shot recognition results.]{Large scale zero-shot recognition results. Flat error in \% and hierarchical error in brackets. $^*$Note that \cite{mensink12eccv,frome13nips} report on a different set of unseen classes than \cite{rohrbach10cvpr}.}
\label{tbl:imgnet:zeroShot}
\end{center}
\end{table} 
 
\citet{rohrbach11cvpr} extend zero-shot recognition from the 10 unseen categories in the AwA dataset to a setting of 200 unseen ImageNet \cite{deng09cvpr} categories. One of the main challenges in this setting is, that there are no pre-defined attributes on this dataset available. \citeauthor{rohrbach11cvpr} propose to mine part-attributes from WordNet \cite{fellbaum:wordnet} as ImageNet categories correspond to WordNet synsets. 
Additionally, as the known and unknown classes are leaf nodes of the ImageNet hierarchy, inner nodes can be used to group leaf nodes, similar to attributes. Also, the closest known leaf node categories can transfer to the corresponding unseen leaf category.

An alternative approach is DeViSE \cite{frome13nips} which learns an embedding into a semantic skip-gram word-space \cite{mikolov13nips}, trained on Wikipedia documents. Classification is achieved by projecting an image in the word-space and taking the closest word as label. Consequently this also allows for zero-shot recognition.

\Tableref{tbl:imgnet:zeroShot} compares the different approaches. The hierarchical variants \cite{rohrbach11cvpr} performs best, also compared to DeViSE \cite{frome13nips} which relies on more powerful CNN \cite{krizhevsky12nips} features. Further improvements can be achieved by metric learning \cite{mensink12eccv}. 
As a different application, \citet{mrowca15iccv} show how such hierarchical semantic knowledge allows to improve large scale object detection not just classification. 
While the WordNet hierarchy is very reliable as it was manually created, the attributes are restricted to part attributes and the mining is not as reliably. To  improve in this challenging setting, we discuss next how one can exploit instance similarity and few labeled examples if available.

\begin{figure}[t]
 \sidecaption[t]
  \includegraphics[width=0.5\linewidth]{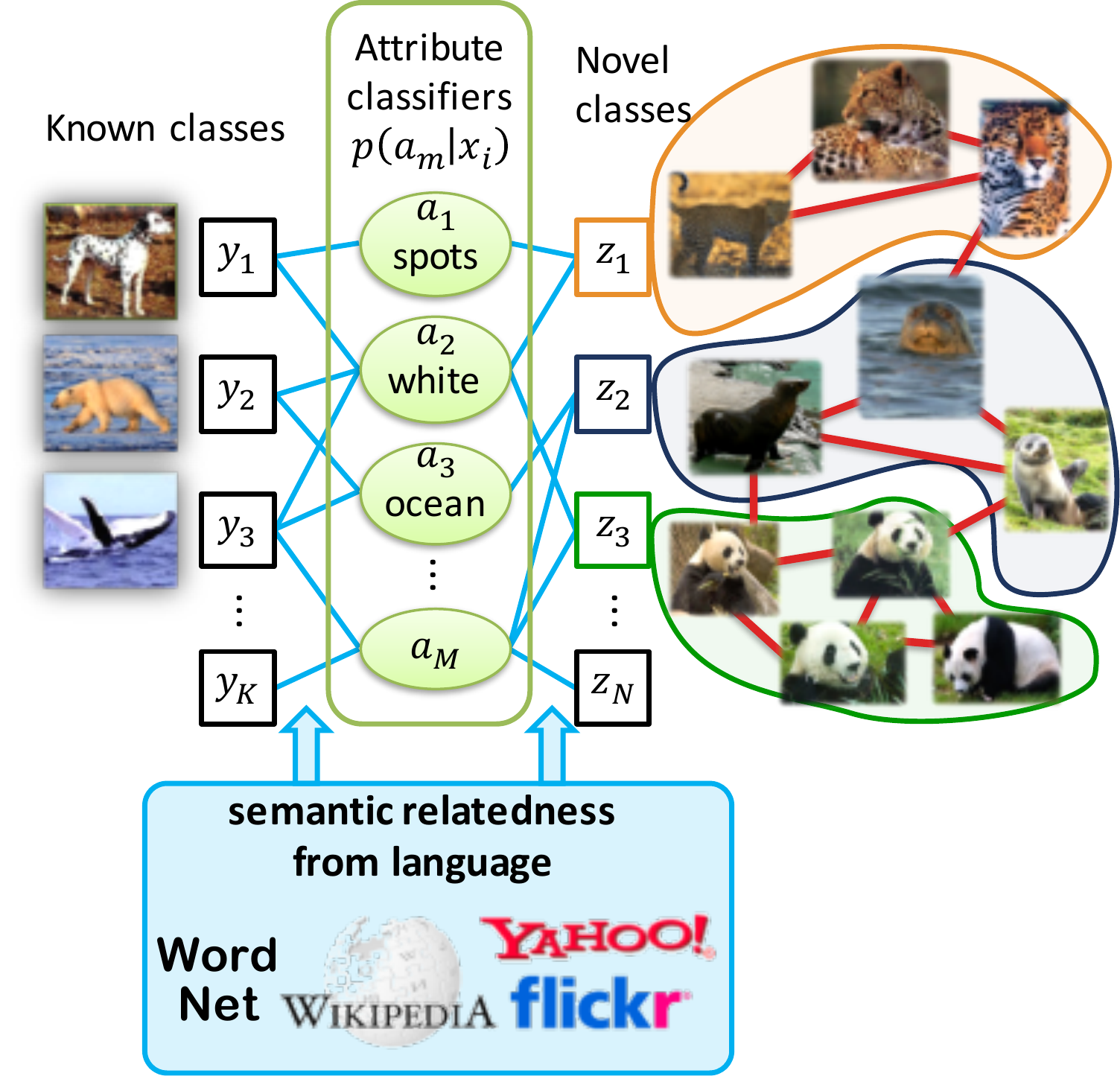}
  \caption{Recognition of novel categories. The approach \emph{Propagated Semantic Transfer} \cite{rohrbach13nips} combines knowledge transferred via attributes from known classes (left) with few labeled examples in graph (red lines) which is build according to instance similarity.}
  \label{fig:pst}
\end{figure}

Transferring knowledge from known categories to novel classes is challenging as it is difficult to estimate visual properties of the novel classes. Approaches discussed in the previous section can not exploit instance similarity or few labeled instances, if available. The approach \emph{Propagated Semantic Transfer} (PST) \cite{rohrbach13nips} combines four ideas to jointly handle the challenging scenario of recognizing novel categories. %
First, PST transfers information from known to novel categories by incorporating external knowledge, such as linguistic or expert-specified information, \eg, by a mid-level layer of semantic attributes as discussed in \Secref{sec:attributes}. 
Second, PST exploits the manifold structure of novel classes similar to unsupervised learning approaches \cite{Weber2000,Sivic2005}. More specifically it adapts the graph-based Label Propagation algorithm \cite{Zhu2003,Zhou2004} -- previously used only for semi-supervised learning \cite{ebert10eccv}~-- to zero-shot and few-shot learning. In this transductive setting information is propagated between instances of the novel classes to get more reliable recognition as visualized with the red graph in \Figref{fig:pst}. Third, PST improves the local neighborhood in such graph structures by replacing the raw feature-based representation with a semantic object- or attribute-based representation. And forth, PST generalizes from zero- to few-shot learning by integrating labeled training examples as certain nodes in its graph based propagation.
Another positive aspect of PST is that attribute or category models do not have to be retrained if novel classes are added which can be an important aspect \eg in a robotic scenario.

\subsection{Propagated semantic transfer}
\label{sec:pst}

\Figref{fig:res:pst:Awa} shows results on the AwA \cite{lampert09cvpr} dataset. We note that in contrast to the previous section the classifiers are trained on all training examples, not only 92 per class.
\Figref{fig:res:pst:Awa}(a) shows zero-shot results, where no training examples are available for the novel or in this case unseen classes. The table compares PST with propagating on a graph based on attribute-classifier similarity versus image descriptor similarity and shows a clear benefit of the former. This variant also outperform DAP and IAP \cite{lampert13pami} as well as Zero-Shot Learning \cite{fu13pami}.
Next we compare PST in the few-shot setting, \ie we add labeled examples per class. In \Figref{fig:res:pst:Awa}(b) we compare PST to two label propagation (LP) baselines \cite{ebert10eccv}.
We first note that PST (red curves) seamlessly moves from zero-shot to few-shot, while traditional LP (blue and black curves) needs at least one training example. 
We first examine the three solid lines. The black curve is the best LP variant from \citet{ebert10eccv} and uses similarity based image features. %
 LP in combination with the similarity metric based on the attribute classifier scores (blue curves) allows to transfer knowledge residing in the classifier trained on the known classes and gives a significant improvement in performance. PST (red curve) additionally transfers labels from the known classes and improves further. %
The dashed lines in \Figref{fig:res:pst:Awa}(b)  provide results for automatically mined associations between attributes and classes from language resources. It is interesting to note that these automatically mined associations achieve performance very close to the manual defined associations (dashed \vs solid). %

\newcommand{\midruleOneTwoZero}{\cmidrule(lr){1-1}  \cmidrule(lr){2-3} }
\begin{figure}[t]
\center
\begin{tabular}{m{0.42\textwidth} m{0.56\textwidth}}  
\begin{tabular}{l r r }
\toprule
  & \multicolumn{2}{c}{Performance} \\ 
Approach & AUC & Acc.\\\midruleOneTwoZero
DAP \citep{lampert13pami} &  81.4 & 41.4 \\
IAP \citep{lampert13pami} &  80.0 & 42.2 \\
Zero-Shot Learning \citep{fu13pami} & n/a & 41.3 \\
PST \cite{rohrbach13nips} \\
\ \ \ \ on image descriptors &  81.2 & 40.5\\ 
\ \ \ \ on attributes &  83.7 & 42.7\\ %
\bottomrule
\end{tabular} & 
\includegraphics[width=0.56\textwidth]{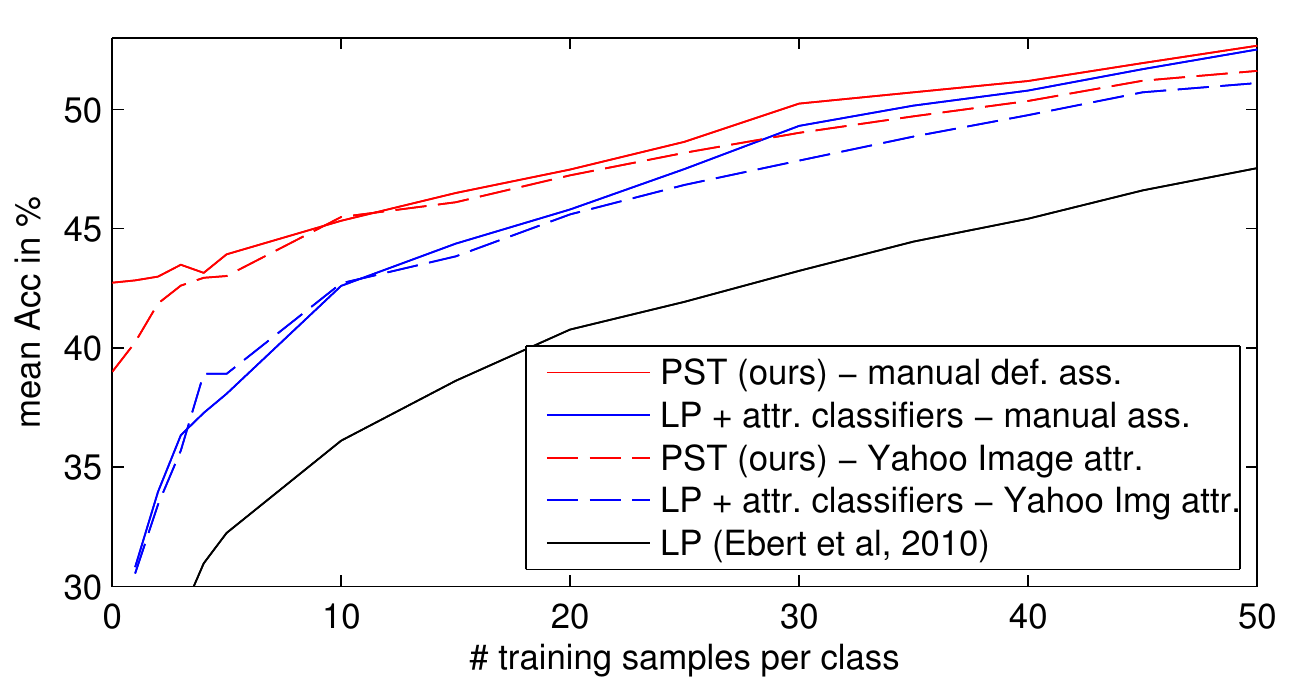} \\
\multicolumn{1}{c}{(a) Zero-Shot, in \%.} & \multicolumn{1}{c}{(b) Few-Shot} 
\end{tabular}
\caption[Zero-shot results on AwA dataset]{Zero-shot results on AwA dataset. Predictions with attributes and manual defined associations. Adapted from \cite{rohrbach13nips}.}
\label{fig:res:pst:Awa}

\end{figure}

\newcommand{\midruleOneOne}{\cmidrule(lr){1-1}  \cmidrule(lr){2-2} }
\begin{figure}[tb]
\center
\begin{tabular}{cc}
\includegraphics[width=0.51\textwidth]{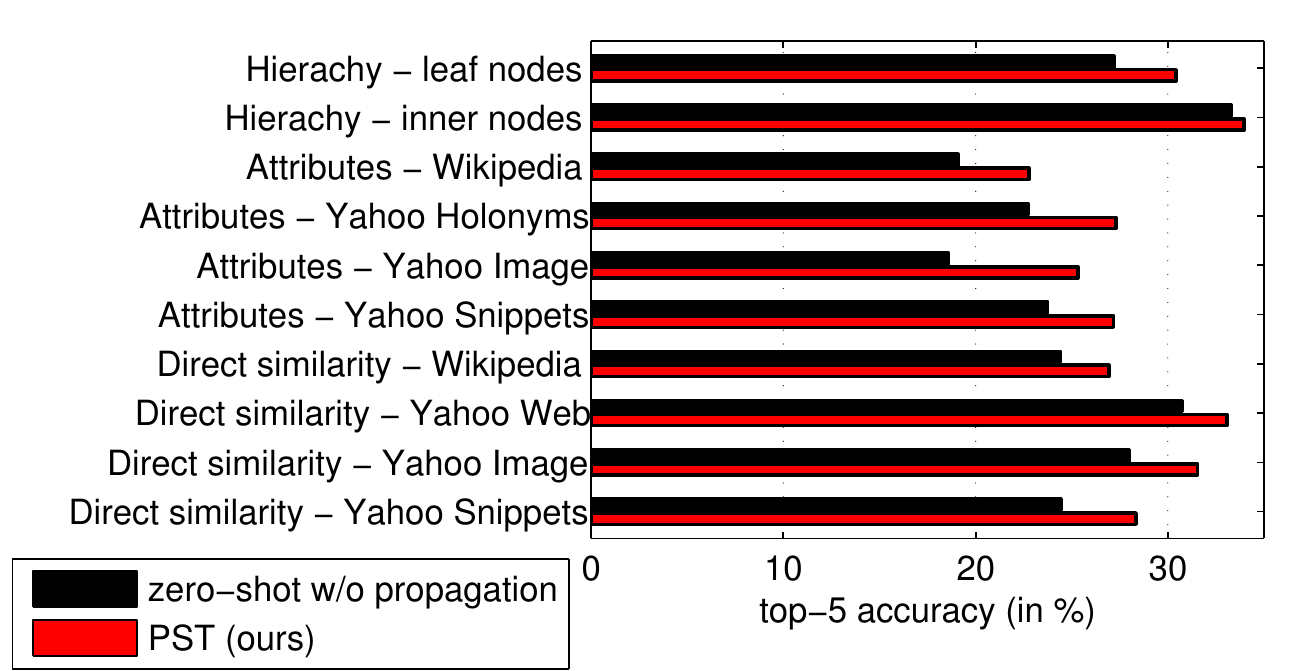}&
\includegraphics[width=0.48\textwidth]{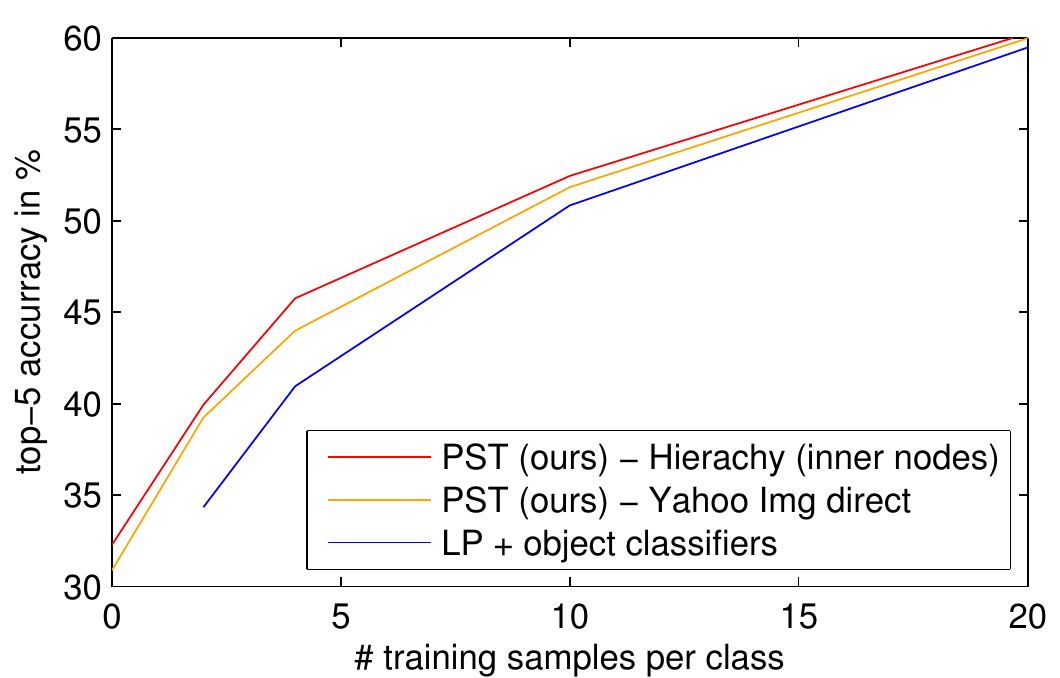}\\
\multicolumn{1}{c}{(a) Zero-Shot recognition} & \multicolumn{1}{c}{(b) Few-Shot recognition} 
\end{tabular}
   \caption[Results on ImageNet]{Results on 200 unseen classes of ImageNet. Adapted from \cite{rohrbach13nips}.}
   \label{fig:nips13:results:ImageNet:ZeroShot}
   \label{fig:nips13:res:ImagenetNshot} 
 \label{fig:nips13:results:ImageNet}
 \end{figure}
\Figref{fig:nips13:results:ImageNet} shows results on the classification task with 200 unseen ImageNet categories. %
In \figref{fig:nips13:results:ImageNet}(a) we compare PST  to zero-shot without propagation presented as discussed in \secref{sec:attributes}. %
For zero-shot recognition PST (red bars) improves performance over zero-shot without propagation (black bars) for all language resources and transfer variants. %
Similar to the AwA dataset, PST also improves over the LP-baseline for few-shot recognition (\figref{fig:nips13:results:ImageNet}b). The missing LP-baseline on raw features is due to the fact that for the large number of images and high dimensional features the graph construction is very time and memory consuming if not infeasible. In contrast, the attribute representation is very compact and thus computational tractable even with a large number of images.

\subsection{Composite activity recognition with attributes and script data}
\label{sec:activityrecognition}

 \begin{figure}[t]
 \sidecaption[t]
  \includegraphics[width=0.64\linewidth]{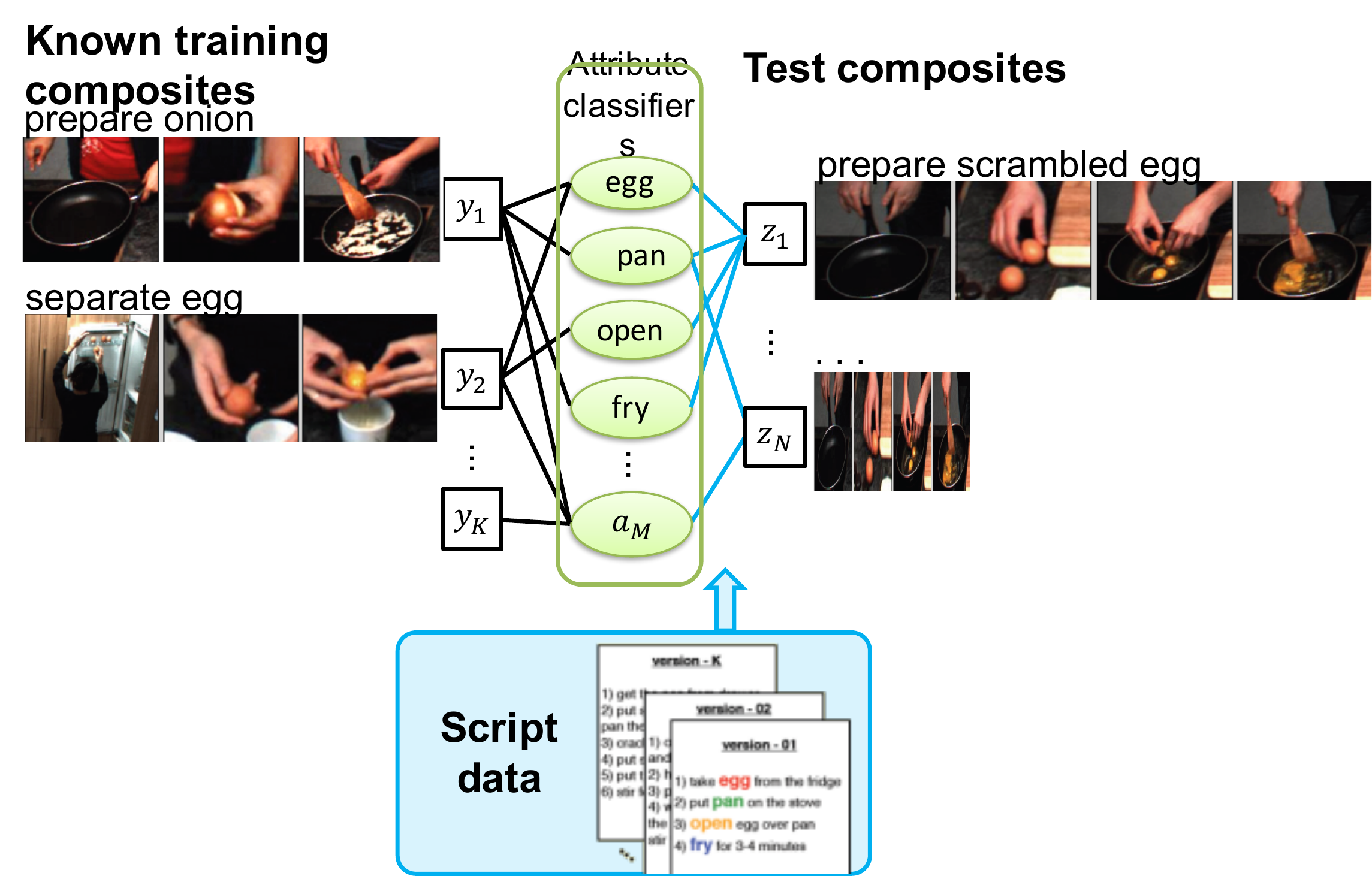}
  \caption{Recognizing composite activities using attributes and script data.}
  \label{fig:compoisteactivites}
\end{figure}

Understanding activities in visual and textual data is generally regarded as more challenging than understanding object categories due to the limited training data, challenges in defining the extend of an activity, and the similarities between activities \cite{regneri13tacl}. However, long-term composite activities can be decomposed in  shorter fine-grained activities \cite{rohrbach12eccv}. Consider for example the composite cooking activities \emph{prepare scrambled egg} which can be decomposed in attributes of fine-grained activities (\eg \emph{open}, \emph{fry}), ingredients (\eg \emph{egg}), and tools (\eg \emph{pan}, \emph{spatula}). These attributes can than be shared and transferred across composite activities as visualized in \Figref{fig:compoisteactivites} using the same approaches as for objects and attributes discussed in the previous section. However, the representations, both on the visual and on the language side have to change. Fine-grained activities and associated attributes are visually characterized by fine-grained body motions and low inter-class variability. In addition to holistic features \cite{wang13iccv}, one consequently should exploit human pose-based \cite{rohrbach12cvpr} and hand-centric \cite{senina14arxiv} features. As the previously discussed language resources do not provide good associations between composite activities and their attributes,  \citet{rohrbach12eccv} collected textual description (\emph{Script data}) of these activities with AMT. From this script data associations can be computed based on either the frequency statistics or, more discriminate, by term frequency times inverse document frequency (\tfidf). %

\Tableref{tbl:results:dish} shows results on the MPII Cooking 2 dataset \cite{rohrbach15ijcv}. Comparing the first column (holistic Dense Trajectory features \cite{wang13iccv}) with the second, shows the benefit of adding the more semantic hand-\cite{senina14arxiv} and pose-\cite{rohrbach12cvpr} features. Comparing line (1) with line (2) or (3) shows the benefit of representing composite activities with attributes as this allows sharing across composite activities. Best performance is achieved with 57.4\% mean AP in line (6) when combining compositional attributes with the Propagated Semantic Transfer (PST) approach (see \secref{sec:pst}) and Script data to determine associations between composites and attributes.

\newcommand{\compositeHeader}{ \toprule\multicolumn{1}{r}{Attribute training on:} & \multicolumn{2}{c}{All} & \multicolumn{2}{c}{Disjoint}\\
&  \multicolumn{2}{c}{Composites} &  \multicolumn{2}{c}{Composites}\\
\cmidrule(lr){1-1} \cmidrule(lr){2-3}  \cmidrule(lr){4-5}
\multicolumn{1}{r}{Activity representation:}& \cite{wang13iccv} & \cite{wang13iccv,senina14arxiv,rohrbach12cvpr}& \cite{wang13iccv} & \cite{wang13iccv,senina14arxiv,rohrbach12cvpr}\\ 
\cmidrule(lr){1-1} \cmidrule(lr){2-3}  \cmidrule(lr){4-5}}
\begin{table}[t]
\begin{center}
\begin{tabular}{l c@{\ \ }c  c@{\ \ }c}
\compositeHeader
\multicolumn{5}{l}{\textbf{With training data for composites}}\\
\multicolumn{5}{l}{\emph{Without attributes}}\\
\ \ \ (1) SVM &    39.8   & 41.1& - & - \\
\multicolumn{5}{l}{\emph{Attributes on gt intervals}}\\
\ \ \ (2) SVM & 43.6   & 52.3& 32.3    &  34.9 \\ %
\multicolumn{5}{l}{\emph{Attributes on automatic segmentation}}\\ 
\ \ \ (3) SVM &                  49.0   & 56.9& 35.7    &   34.8\\%SVM - SgemzScore01Train
\ \ \ (4) NN &                   42.1   & 43.3& 24.7   &  32.7 \\%NN on normalized Segm datazScore01Test
\ \ \ (5) NN+\Scriptknowledge&   35.0   & 40.4& 18.0    &21.9        \\%NN+scripts:tfidf-WN
\ \ \ (6) PST+\Scriptknowledge&  54.5   & 57.4& 32.2  &32.5       \\ 
\cmidrule(lr){1-1} \cmidrule(lr){2-3}  \cmidrule(lr){4-5}
\multicolumn{5}{l}{\textbf{No training data for composites }}\\
\multicolumn{5}{l}{\emph{Attributes on automatic segmentation}}\\
\ \ \ (7) \Scriptknowledge& 36.7        & 29.9 & 19.6 &21.9\\
\ \ \ (8) PST + \Scriptknowledge& 36.6    & 43.8 & 21.1&19.3 \\
\bottomrule \\ \end{tabular}
\caption[Composite cooking activity classification.]{Composite cooking activity classification on MPII Cooking 2 \cite{rohrbach15ijcv}, mean AP in \%. Top left quarter: fully supervised, right column: reduced attribute training data, bottom section: no composite cooking activity training data, right bottom quarter: true zero shot.  Adapted from \cite{rohrbach15ijcv}.}
\label{tbl:results:dish}
\label{tbl:results:tasks}
\end{center}
\end{table}

%% file: description.tex
\section{Image and video description using compositional attributes}
\label{sec:description}
In this section we discuss how we can generate natural language sentences describing visual content, rather than just giving labels to images and videos as discussed in the previous section. 
This intriguing task has recently received increased attention in computer vision and computational linguistics communities \cite{venugopalan15iccv,venugopalan15naacl,vinyals15cvpr} and has a large number of potential applications including human robot interaction, image and video retrieval, and describing visual content for visually impaired people. 
In this section we focus on approaches which decouple the visual recognition and the sentence generation and introduce an intermediate semantic layer, which can be seen a layer of attributes (\secref{sec:translate}). Introducing such a semantic layer has several advantages. First, this allows to reason across sentences on a semantic level, which is, as we will see, beneficial for multi-sentence description of videos (\secref{sec:multisentence}).  Second, we can show that when learning reliable attributes, this leads to state-of-the-art sentences generation with high diversity in the challenging scenario of movie description (\secref{sec:moviedescription}). Third, this leads to a compositional structure which allows describing novel concepts in images and videos (\secref{sec:novel_sentences}).

\subsection{Translating image and video content to natural language descriptions}

\begin{figure}[t]
\begin{center}
  \includegraphics[width=\textwidth]{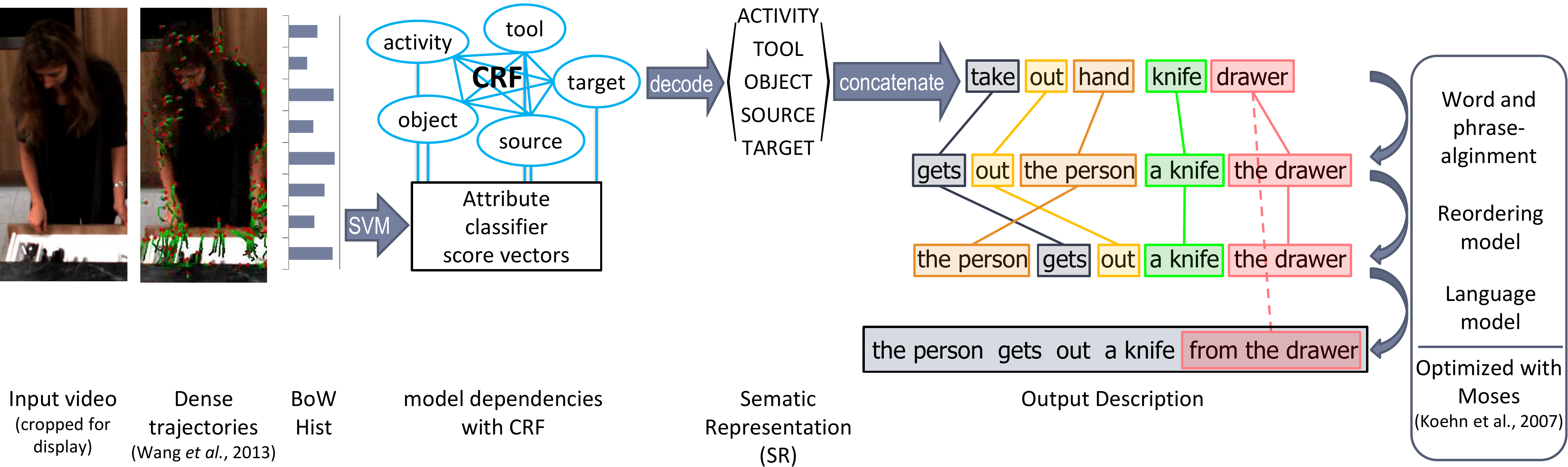}
  \caption{Video description. Overview of the two-step translation approach \cite{rohrbach13iccv} with an intermediate semantic layer of attributes (SR) for describing videos with natural language. From \cite{rohrbach14phd}.}   \label{fig:iccv13:approach}
\end{center}
\end{figure}

To address the problem of image and video description, \citet{rohrbach13iccv} propose a two-step translation approach which first predicts an intermediate semantic attribute layer and then learns how to translate from this semantic representation to natural sentences. %
Figure \ref{fig:iccv13:approach}  gives an overview of this two-step approach for videos.
First, a rich semantic representation of the visual content including \eg object and activity attributes is predicted. To predict the semantic representation a CRF models the relationships between different attributes of the visual input. And second, the generation of natural language is formulated as a machine translation problem using the semantic representation as source language and the generated sentences as target language.  For this a parallel corpus of videos, annotated semantic attributes, and textual descriptions allows to adapt statistical machine translation (SMT) \cite{koehn10book} to translate between the two languages. \citeauthor{rohrbach13iccv} train and evaluate their approach on the videos of the MPII Cooking dataset \cite{rohrbach12cvpr,rohrbach12eccv} and the aligned descriptions from the TACoS corpus \cite{regneri13tacl}.
According to automatic evaluation and human judgments, the two-step translation approach significantly outperforms retrieval and n-gram-based baseline approaches, motivated by prior work. This similarly can be applied to image description task, however, in both cases it requires an annotated semantic attribute representation. In \secsref{sec:moviedescription} and \ref{sec:novel_sentences} we discuss how we can extract such attribute annotations automatically from sentences.  An alternative approach is presented by \citet{fang15cvpr} who mine visual concepts for image description by integrating multiple instance learning \cite{maron1998framework}. Similar to the work presented in the following, \citet{wu16arxiv1603.02814} learn an intermediate attribute representation from the image descriptions. Captions are then generated solely from the intermediate attribute representation.

\label{sec:translate}

\subsection{Coherent multi-sentence video description with variable level of detail}

\begin{figure}[tp]
\begin{center}
\includegraphics[scale=0.32]{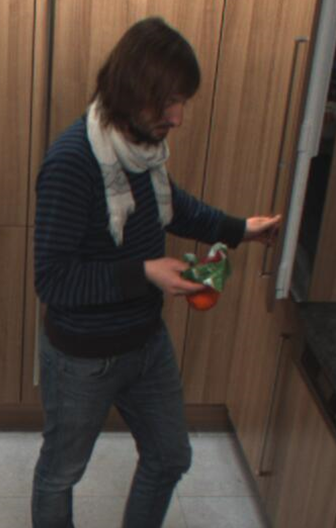}
\includegraphics[scale=0.32]{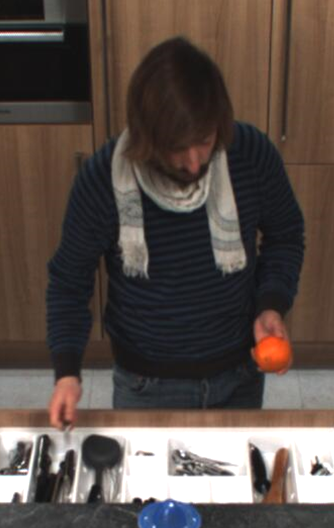}
\includegraphics[scale=0.32]{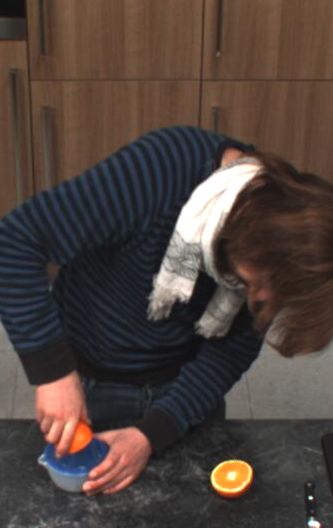}
\includegraphics[scale=0.32]{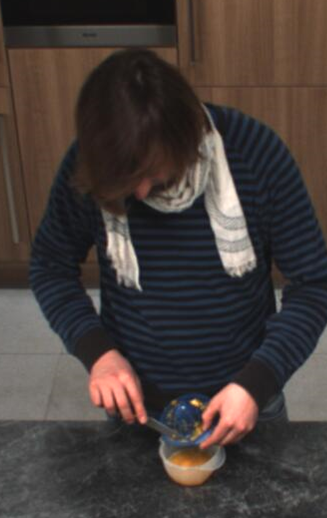}
\includegraphics[scale=0.32]{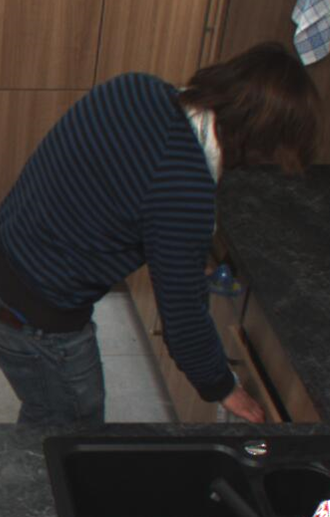}
\includegraphics[scale=0.32]{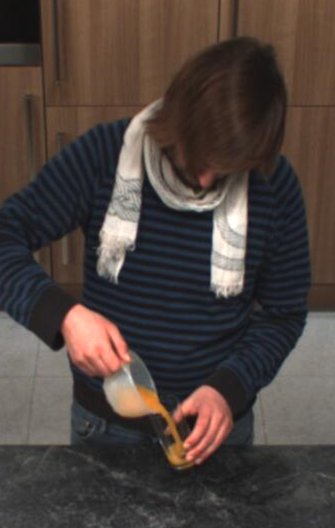}\\
\end{center}
\begin{tabular}{@{}p{1.8cm}p{9.4cm}@{}}
\textbf{Detailed:}&
A man took a cutting board and knife from the drawer. He took out an orange from the refrigerator. Then, he took a knife from the drawer. He juiced one half of the orange. Next, he opened the refrigerator. He cut the orange with the knife. The man threw away the skin. He got a glass from the cabinet. Then, he poured the juice into the glass. Finally, he placed the orange in the sink.\\
\textbf{Short:}&
A man juiced the orange. Next, he cut the orange in half. Finally, he poured the juice into a glass.\\
\textbf{One sentence:}&
A man juiced the orange.\\
\end{tabular}
\caption{Coherent multi-sentence descriptions at three levels of detail, using automatic temporal segmentation. See \secref{sec:multisentence} for details. From \cite{rohrbach14gcpr}.}
\label{fig:gcpr14}
\end{figure}

Most approaches for automatic video description, including the one presented above, focus on generating single sentence descriptions and are not able to vary the descriptions' level of detail. 
One advantage of the two-step approach with an explicit intermediate layer of semantic attributes is that it allows to reason on this semantic level.  
To generate coherent multi-sentence descriptions, \citet{rohrbach14gcpr} extend the two-step translation approach to model across-sentence consistency at the semantic level by enforcing a consistent topic, which is the prepared dish in the cooking scenario. To produce shorter or one-sentence summaries, \citeauthor{rohrbach14gcpr} select the most relevant sentences on the semantic level by using \tfidf (term frequency times inverse document frequency). For an example output on the TACoS Multi-Level corpus  \cite{rohrbach14gcpr} see Figure \ref{fig:gcpr14}.
In order to fully automatically do multi-sentence description, \citeauthor{rohrbach14gcpr} propose a simple but effective method based on agglomerative clustering to perform automatic video segmentation. The most important component of good clustering is the similarity measure and it turns out that the semantic attribute classifiers (see \Figref{fig:iccv13:approach}) are very well suited for that in contrast to Bag-of-Words dense trajectories \cite{wang11cvpr}. This confirm the observation made in \secref{sec:pst} that attribute classifiers seem to form a good space for distance computations.

To improve performance, \citet{donahue15cvpr} show that the second step, the SMT-based sentence generation, can be replaced with a deep recurrent network to better model visual uncertainty, but still relying on the multi-sentence reasoning on the semantic level. On the TACoS Multi-Level corpus this achieves 28.8\% BLEU@4, compared to 26.9\% \cite{rohrbach14gcpr} with SMT and 24.9\%  with SMT without multi-sentence reasoning \cite{rohrbach13iccv}.

\label{sec:multisentence}

\subsection{Describing movies with an intermediate layer of attributes}
\label{sec:visionlang}
\label{sec:moviedescription}
Two challenges arise, when extending the  idea presented above to movie description \cite{rohrbach15cvpr}, which looks at the problem how to describe movies for blind people. First, and maybe more importantly, there are no semantic attributes annotated as on the kitchen data, and second, the data is more visually diverse and challenging.
For the first challenge, \citet{rohrbach15cvpr} propose to extract attribute labels from the description to train visual classifiers to build a semantic intermediate layer by relying on a semantic parsing approach of the description. To additionally accommodate the second challenge of increased visual difficulty, \citet{rohrbach15gcpr} show how to improve the robustness of these attributes or ``Visual Labels'' by three steps. First, by distinguishing three semantic groups of labels (verbs, objects and scenes) and using corresponding feature representations for each: activity recognition with dense trajectories \cite{wang13iccv}, object detection with LSDA \cite{hoffman14nips}, and scene classification with Places-CNN \cite{zhou14nips}. Second, training  each semantic group separately, which removes noisy negatives. And third, selecting only the most reliable classifiers.
While \citeauthor{rohrbach15cvpr} use SMT for sentence generation in  \cite{rohrbach15cvpr}, they rely on a recurrent network (LSTM) in \cite{rohrbach15gcpr}.

\input{15gcprqualresults}
 The Visual Labels approach outperforms prior work \cite{venugopalan15arxiv1505.00487v2,rohrbach15cvpr,yao15arxiv} on the MPII-MD \cite{rohrbach15cvpr} and M-VAD \cite{torabi15arxiv} dataset with respect to automatic and human evaluation. Qualitative results are shown in \Figref{fig:gcpr15:qual}. An interesting characteristic of the compared methods is the size of the output vocabulary, which is \emph{94} for \cite{rohrbach15cvpr}, \emph{86} for \cite{venugopalan15arxiv1505.00487v2} (which uses an end-to-end LSTM approach without an intermediate semantic representation) and \emph{605} for \cite{rohrbach15gcpr}. Although it is far lower than \emph{6,422} for the human reference sentences, it clearly shows a higher diversity of the output for \cite{rohrbach15gcpr}. %

%% file: 15gcprqualresults.tex
\begin{figure}[t]
\scriptsize
\center
\begin{tabular}{@{}l@{\ \ \ }l@{\ \ \ }l@{}}
\toprule
\multirow{5}{*}{\includegraphics[width=2.5cm]{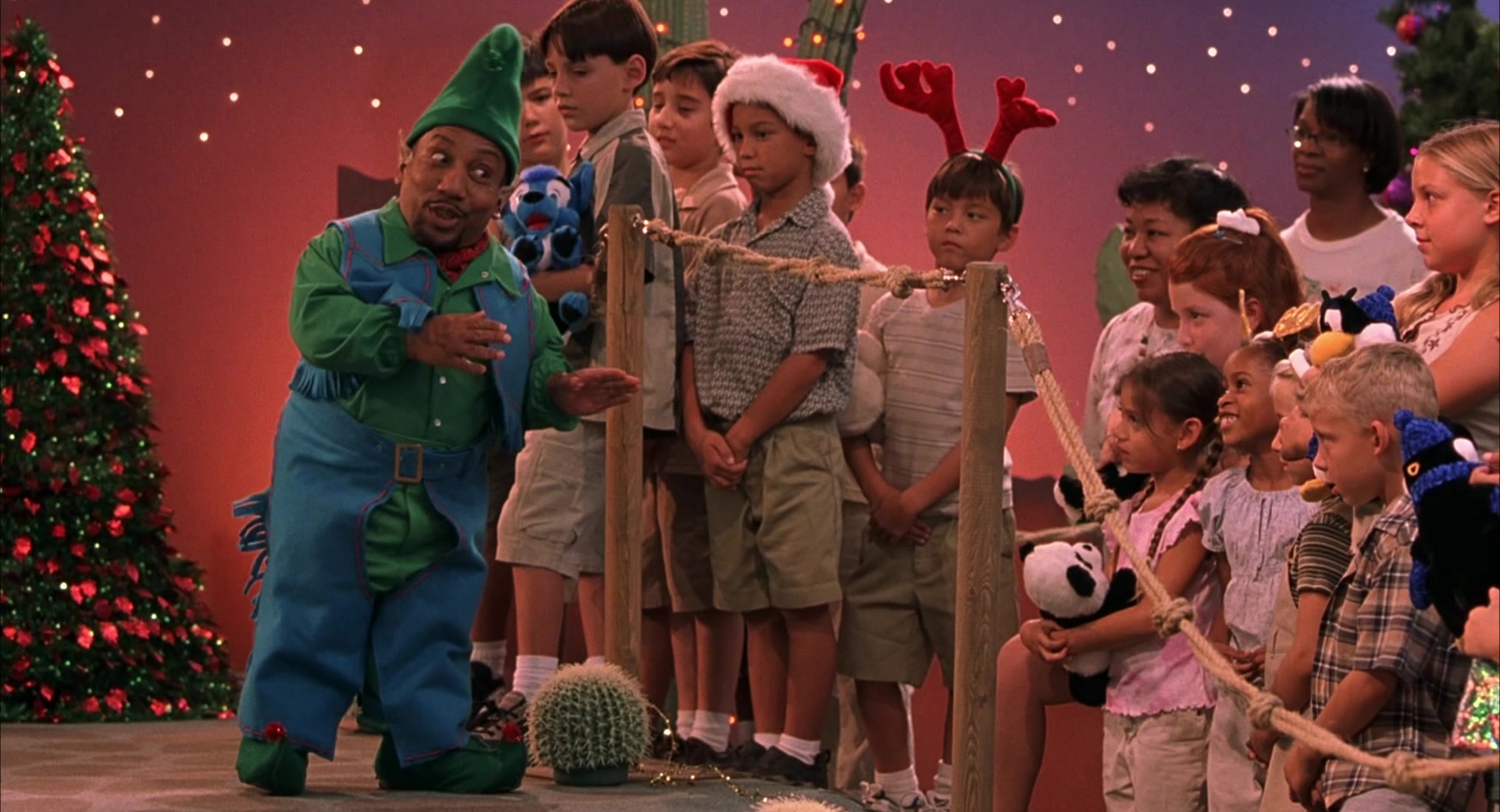}} & SMT \cite{rohrbach15cvpr} & Someone is a man, someone is a man. \\ 
 &S2VT \cite{venugopalan15arxiv1505.00487v2} & Someone looks at him, someone turns to someone.\\ 
 &Visual labels \cite{rohrbach15gcpr} & Someone is standing in the crowd, a little man with a little smile. \\
 & Reference & Someone, back in elf guise, is trying to calm the kids. \\
 \\
\multirow{4}{*}{\includegraphics[width=2.5cm]{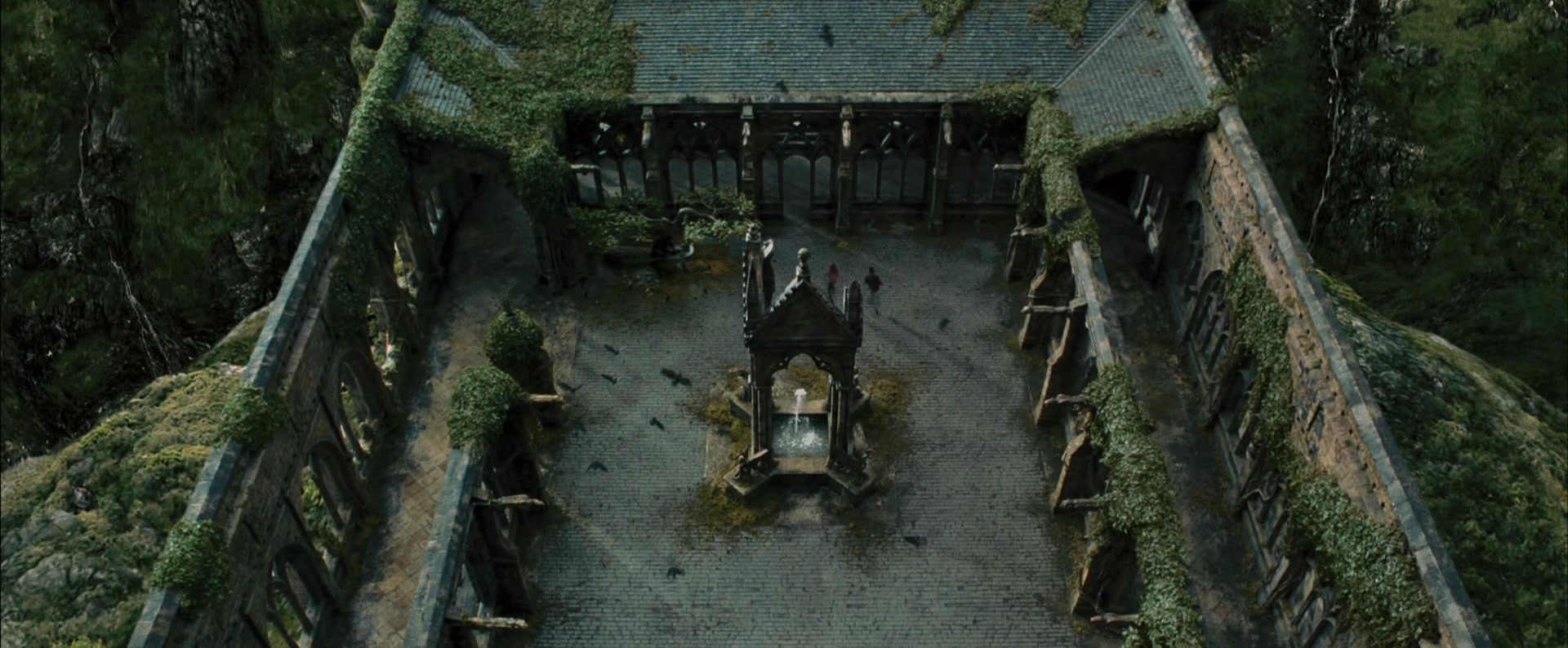}} & SMT \cite{rohrbach15cvpr} &The car is a water of the water. \\
 &S2VT \cite{venugopalan15arxiv1505.00487v2} &  On the door, opens the door opens. \\
  &Visual labels \cite{rohrbach15gcpr} &  The fellowship are in the courtyard. \\
  &Reference & They cross the quadrangle below and run along the cloister. \\
  \\
\multirow{5}{*}{\includegraphics[width=2.5cm]{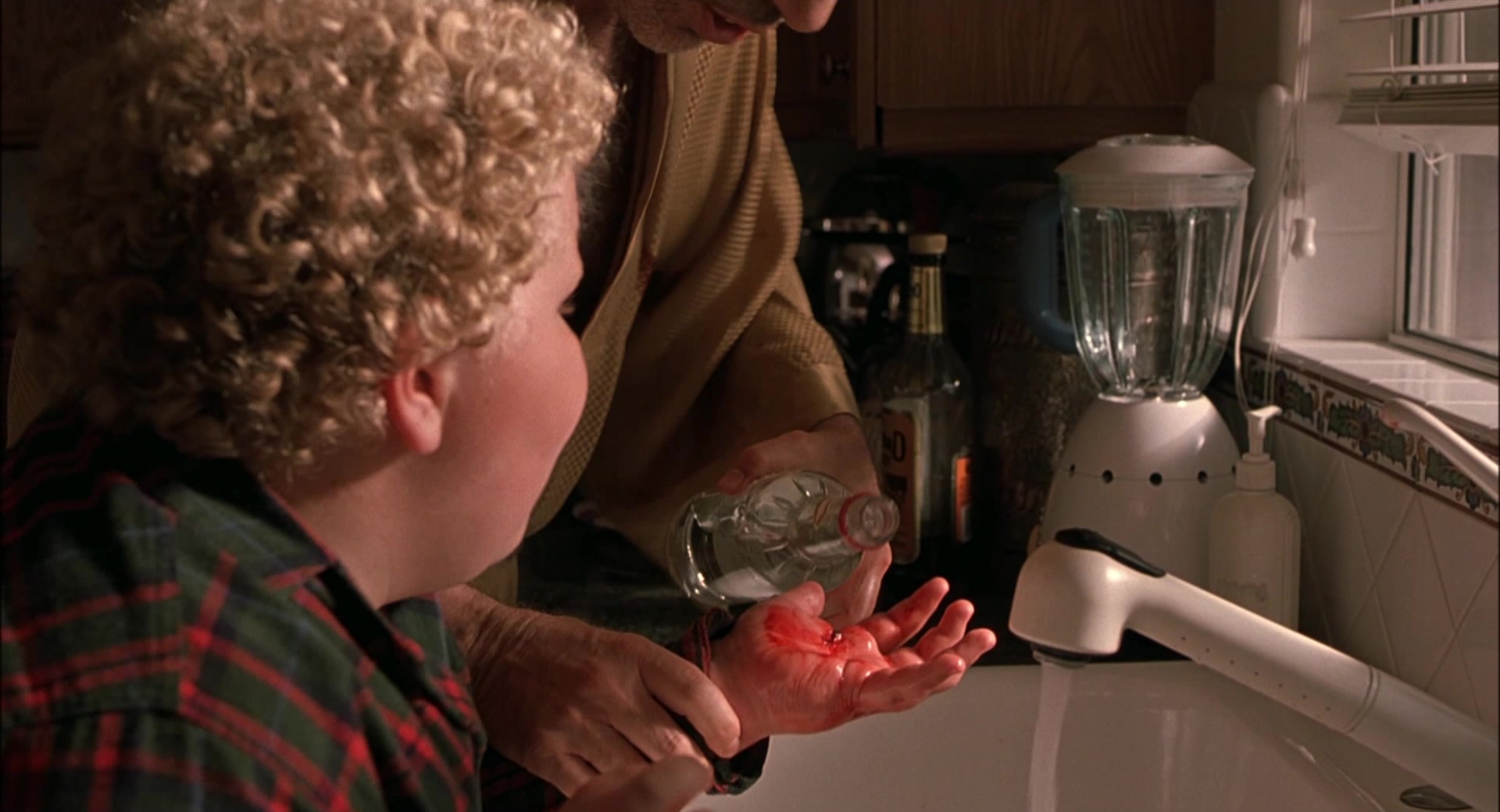}}& SMT \cite{rohrbach15cvpr} & Someone is down the door, someone is a back of the door,\\
& & and someone is a door. \\
 &S2VT \cite{venugopalan15arxiv1505.00487v2} &   Someone shakes his head and looks at someone.\\ 
 &Visual labels \cite{rohrbach15gcpr} & Someone takes a drink and pours it into the water. \\
 &Reference& Someone grabs a vodka bottle standing open on the counter \\
 && and liberally pours some on the hand. \\
\bottomrule
\end{tabular}
\vspace{-0.2cm}
\caption{Qualitative results on the MPII Movie Description (MPII-MD) dataset \cite{rohrbach15cvpr}. The ``Visual labels'' approach \cite{rohrbach15gcpr} which uses an intermediate layer of robust attributes, identifies activities, objects, and places better than related work. From \cite{rohrbach15gcpr}.}
\label{fig:gcpr15:qual}
\end{figure}

%% file: novelimgdesc.tex
\subsection{Describing novel object categories}
\label{sec:novel_sentences}

\begin{figure}[t]
\sidecaption[t]
\includegraphics[width=.64\linewidth]{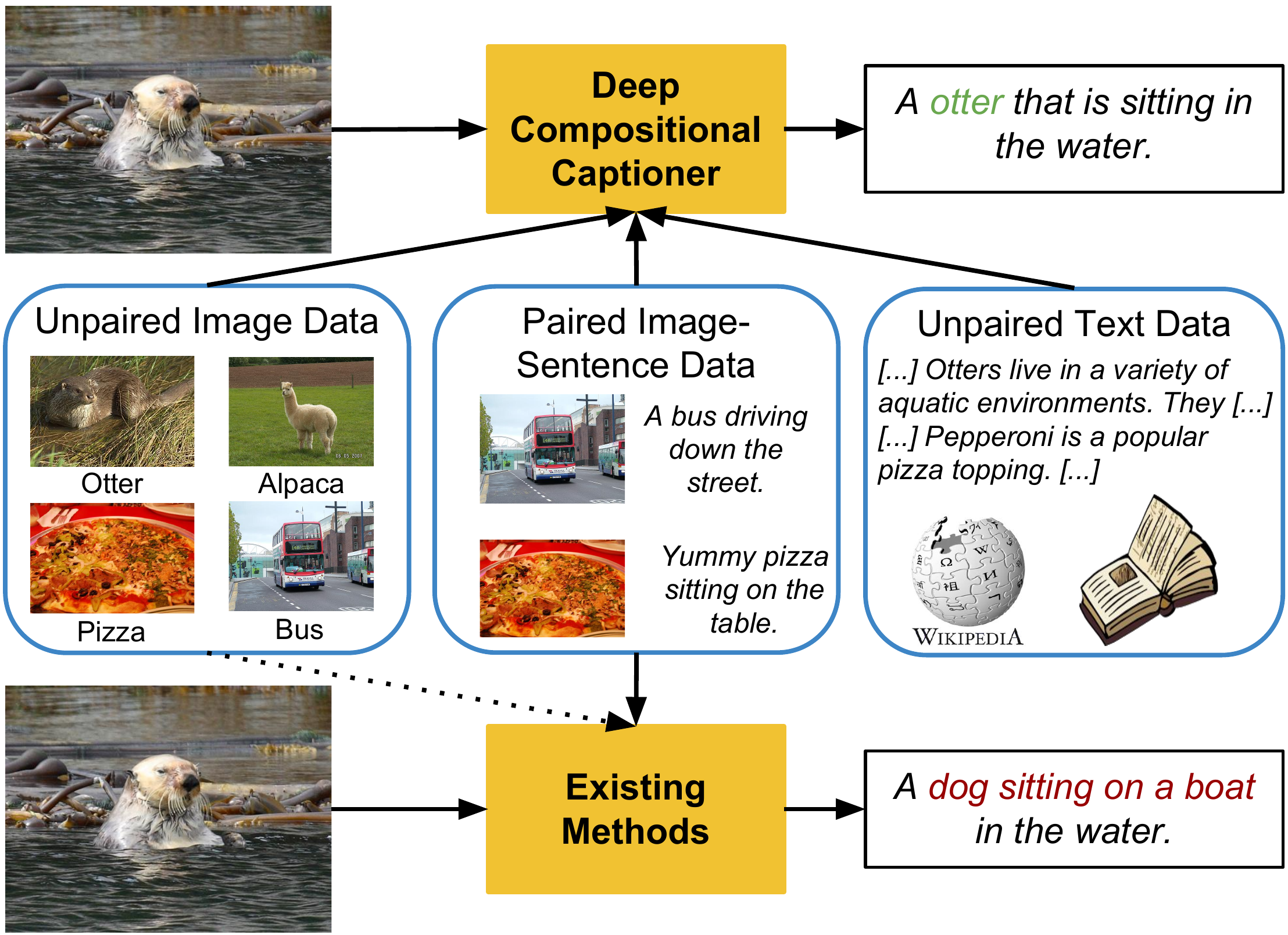}
   \caption{Describing novel object categories which are not contained in caption corpora (like otter). The Deep Compositional Captioner (DCC) \cite{hendricks16cvpr} uses an intermediate semantic attribute or ``lexical'' layer to connect classifiers learned on unpaired image datasets (ImageNet) with text corpora (\eg Wikipedia). This allows it to compose descriptions about novel objects without any paired image-sentences training data.  Adapted from \cite{hendricks15arxiv}.}
\label{fig:novelObjects}
\end{figure}

In this section we discuss how to describe novel object categories which combines challenges discussed for recognizing novel categories (\secref{sec:objectrecognition}) and generating descriptions (\secref{sec:translate}).
State-of-the-art deep image and video captioning approaches (\eg \cite{vinyals15cvpr,mao15iclr,donahue15cvpr,fang15cvpr,venugopalan15iccv}) are limited to describe objects which appear in caption corpora such as MS COCO \cite{chen15arXiv1504.00325} which consist of pairs of images and sentences. In contrast, labeled image datasets without sentence descriptions (\eg ImageNet \cite{deng10eccv}) or text only corpora (\eg Wikipedia) cover many more object categories. %

\citet{hendricks16cvpr} propose the Deep Compositional Captioner (DCC) to exploit these vision-only and language-only unpaired data sources to describe novel categories as visualized in \Figref{fig:novelObjects}. Similar to the attribute layer discussed in \secref{sec:translate}, \citeauthor{hendricks16cvpr} extract words as labels from the descriptions to learn  a ``Lexical Layer''. The Lexical Layer is expanded by objects from ImageNet \cite{deng10eccv}. To be able to not only recognize but also generate the description about the novel objects, DCC transfers the word prediction model from semantically closest known word in the Lexical Layer, where similarity is computed with Word2Vec \cite{mikolov13nips}. %
Interesting to note is, that image captioning approaches such as \cite{vinyals15cvpr,donahue15cvpr} do use ImageNet data to (pre-) train the models (indicated with a dashed arrow in \Figref{fig:novelObjects}), but they do not make use of the semantic information but only the learned representation. 

\begin{figure}[t]
\hspace{-1cm}
\includegraphics[scale=0.7]{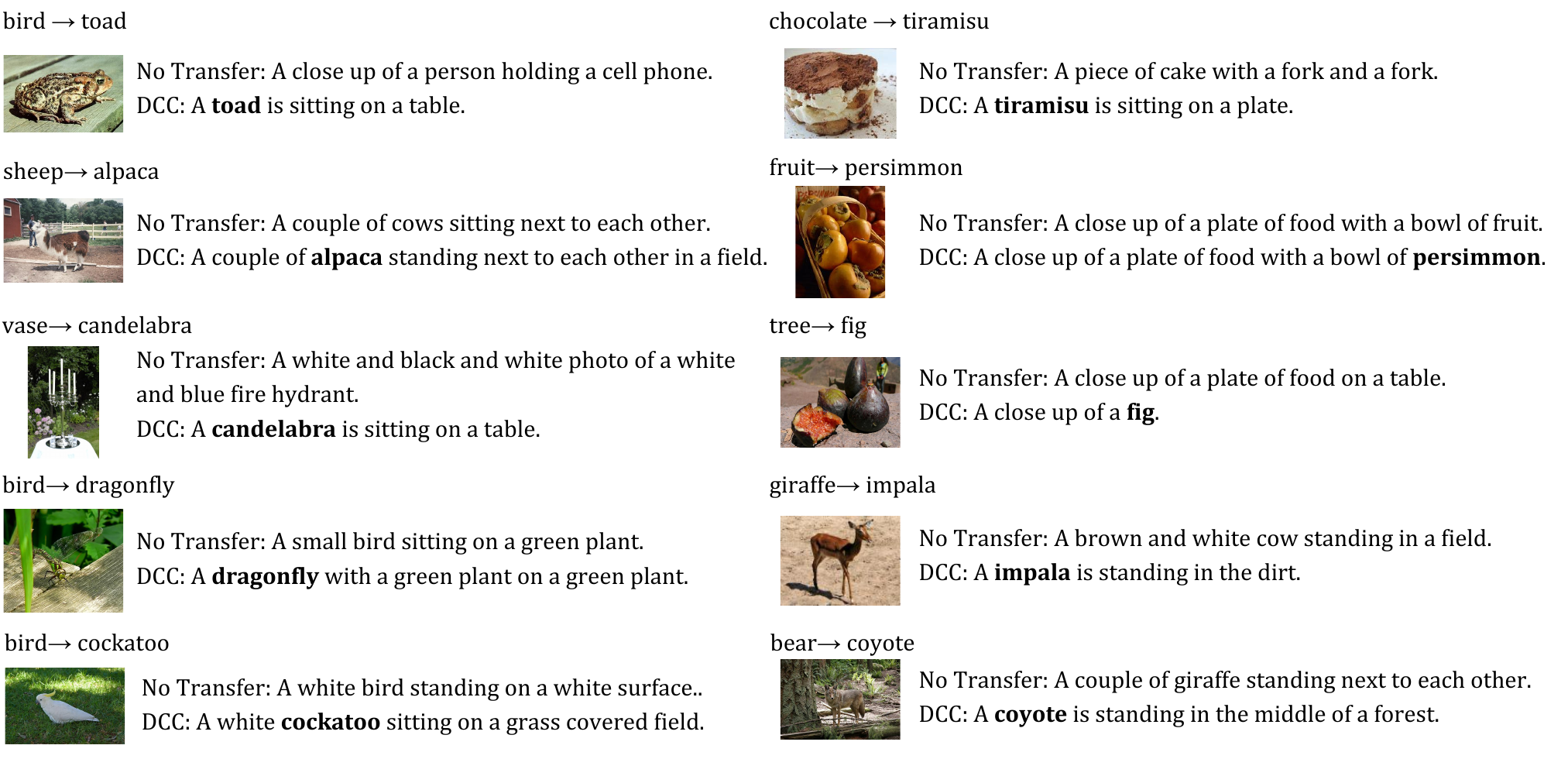}
    \caption{Qualitative results for describing novel ImageNet object categories.  DCC \cite{hendricks16cvpr} compared to an ablation without transfer. X $\rightarrow$ Y: known word X is transferred to novel word Y. From \cite{hendricks15arxiv}.}
    \label{fig:description:res:qual}
\end{figure}

\Figref{fig:description:res:qual} shows several categories where there exist no captions for training. With respect to quantitative measures, compared to a baseline without transfer, DCC improves METEOR from 18.2\% to 19.1\% and F1 score, which measures the appearance of the novel object, from 0 to 34.3\%. \citeauthor{hendricks16cvpr} also show  similar results for video description.

%% file: grounding.tex
\section{Grounding text in images}
\label{sec:ground}

\begin{figure}[t]
\hspace{-1cm}
\begin{tabular}{p{0.48\textwidth}p{0.34\textwidth}p{0.27\textwidth}}
\ \ \includegraphics[width=\linewidth]{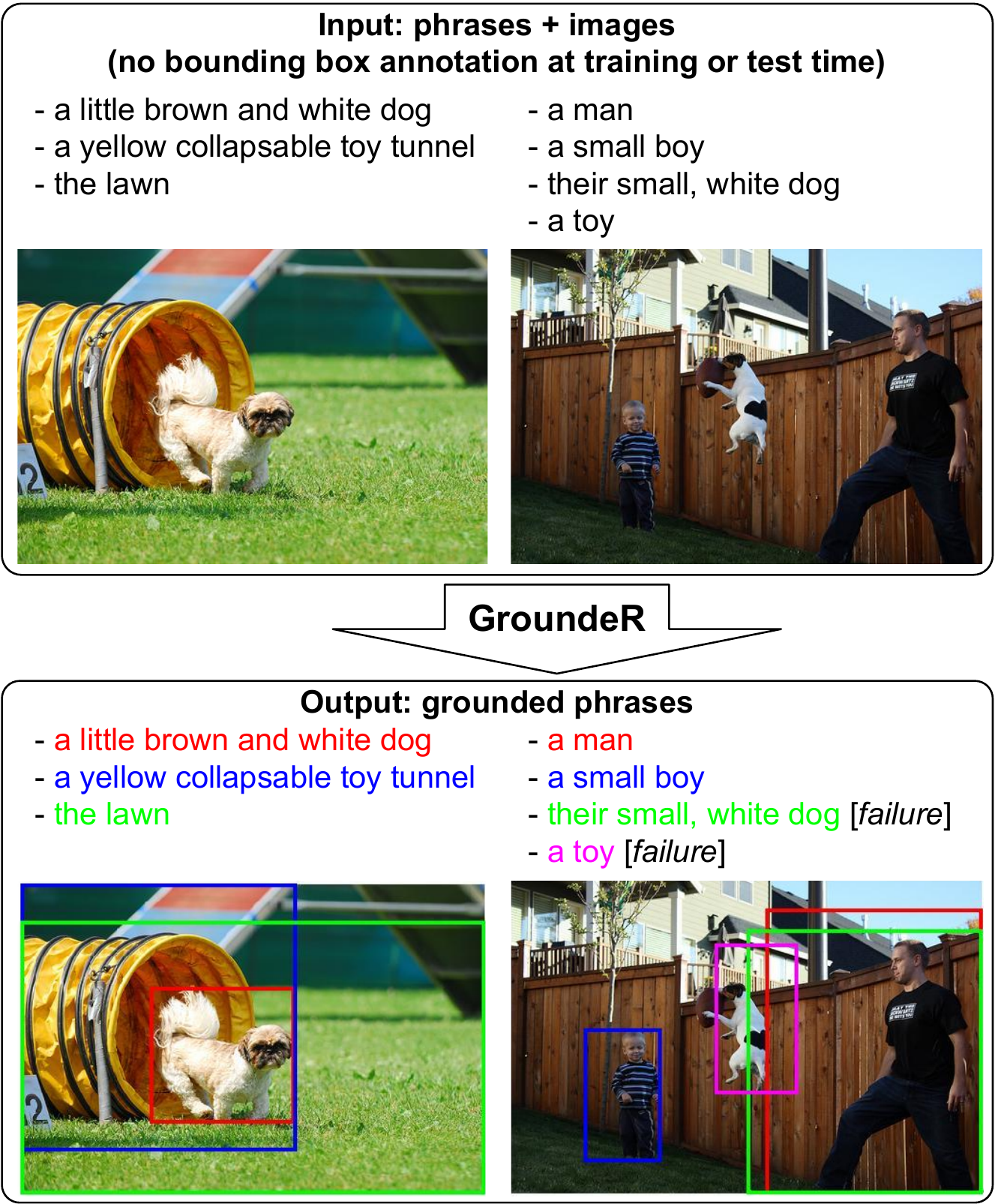}& 
\includegraphics[width=\linewidth]{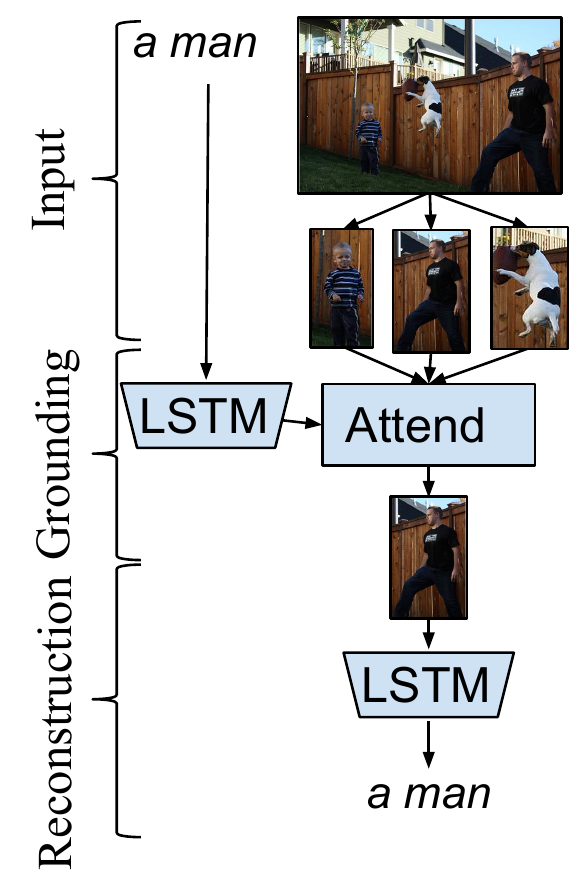} & 
\includegraphics[width=\linewidth]{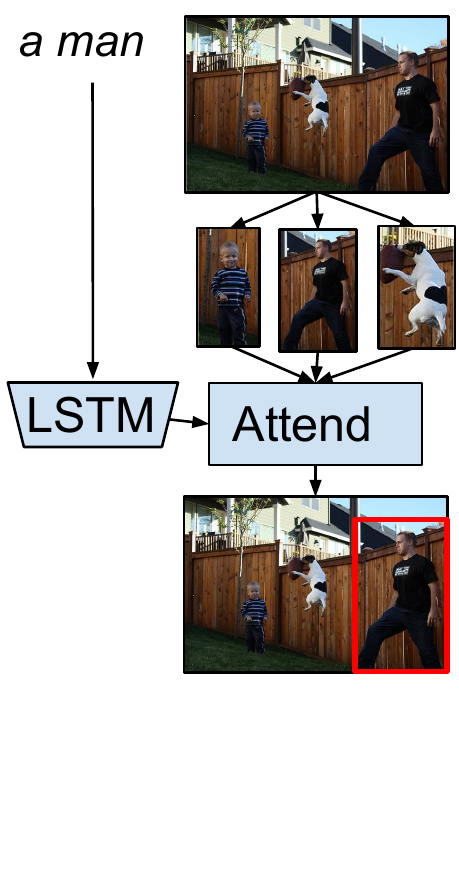}\\
(a) Without bounding box annotations at training or test time \approach \cite{rohrbach15arxiv1511.03745} learns to ground free-form natural language phrases in images.& (b) \approach \cite{rohrbach15arxiv1511.03745} reconstructs phrases by learning to attend to the right box at training time. &
(c)  \approach \cite{rohrbach15arxiv1511.03745} localizes boxes test time.\\ 
   \end{tabular}
   \caption{Unsupervised grounding by learning to associate visual and textual semantic units. From \cite{rohrbach15arxiv1511.03745}.}
\label{fig:ground:unsupervised:teaser}
\label{fig:concept}
\end{figure}

In this section we discuss the problem of grounding natural language in images. Grounding in this case means that given an image and a natural language sentence or phrase, we aim to localize the subset of the image which corresponds to the input phrase. For example, for the sentence ``\emph{A little brown and white dog
emerges from a yellow collapsable
toy tunnel onto the
lawn.}'' and the corresponding image in \Figref{fig:ground:unsupervised:teaser}(a), we want to segment the sentence into phrases and locate the corresponding bounding boxes (or segments) in the image.
While grounding has been addressed \eg in 
\cite{kong14cvpr,johnson2015cvpr,barnard03jmlr,socher10cvpr}, it is restricted to few categories. %
An exception are \citeauthor{karpathy14nips} \cite{karpathy15cvpr,karpathy14nips} who aim to discover a latent alignment between phrases in text and bounding box proposals in the image. \citet{karpathy14nips} ground dependency-tree relations to image regions using multiple instance learning (MIL) and a ranking objective. \citet{karpathy15cvpr} simplify the MIL objective to just the maximal scoring box and replace the dependency tree with a learned recurrent network.
These approaches have unfortunately not been evaluated with respect to the grounding performance due to a lack of annotated datasets. Only recently two datasets were released: Flickr30k Entities \cite{plummer15iccv} augments Flickr30k  \cite{young2014image} with bounding boxes for all noun phrases present in textual descriptions and ReferItGame \cite{kazemzadeh14emnlp} has localized referential expressions in images. Even more recent, at the time of writing, efforts are being made to also collect grounded referential expressions for the MS COCO \cite{coco2014} dataset, namely the authors of ReferItGame are in progress of extending their annotations as well as longer referential expressions have been collected by \citet{mao16cvpr}.
Similar efforts are also made in the Visual Genome project \cite{krishnavisualgenome} which provides densely annotated images with phrases.

In the following we focus on how to approach this problem and the first question is, where is the best point of interaction between linguistic elements and visual elements?
Following the approaches in \cite{karpathy15cvpr,karpathy14nips,plummer15iccv} a good way to this is to decompose both, sentence and image into concise semantic units or attributes which we can match to each other.
For the data as shown in  \Figsref{fig:ground:unsupervised:teaser}(a) and \ref{fig:flickr-unsupervised}, sentences can be split into phrases of typically a few words and images are composed into a larger number of bounding box proposals \cite{uijlings2013selective}. %
An alternative is to integrate  phrase grounding in a fully-convolutional network, for bounding box prediction \cite{johnson16cvpr} or segmentation prediction \cite{hu16arxiv1603.06180}.
In the following, we discuss approaches which focus on how to find the association between visual and linguistic components, rather than the actual segmentation into components. 
We first look at an unsupervised setting with respect to the grounding task, \ie we assume that no bounding box annotations are available for training (\Secref{sec:ground:unsupervised}), and then we show how to integrate supervision (\Secref{sec:grounding:supervised}). \Secref{sec:grounding:results} discusses the results.

\subsection{Unsupervised grounding}
\label{sec:ground:unsupervised}

\input{grounding_examples}

 Although many data sources contain images which are described with sentences or phrases, they typically do not provide the spatial localization of the phrases. This is true for both curated datasets such as MSCOCO \cite{coco2014} or large user generated content as \eg in the YFCC 100M dataset \cite{thomee2016yfcc100m}.
Consequently, being able to learn from this data without grounding supervision would allow large amount and variety of training data. This setting is visualized in \Figref{fig:ground:unsupervised:teaser}(a).

For this setting \citet{rohrbach15arxiv1511.03745} propose the approach \approach, which is able to learn the grounding by aiming to reconstruct a given phrase using an attention mechanism as shown in \Figref{fig:concept}(b). In more detail, given images paired with natural language phrases (or sentence descriptions), but without any bounding box information, we want to localize these phrases with a bounding box in the image (\Figref{fig:concept}c). To do this, \approach learns to attend to a bounding box proposal and, based on the selected bounding box, reconstructs the phrase (\Figref{fig:concept}b). Attention means that the model predicts a weighting over the bounding boxes and then takes the weighted average of the features from all boxes. A softmax over the weights encourages that only one or a few boxes have high weights. As the second part of the model (\Figref{fig:concept}b, bottom) is able to predict the correct phrase only if the first part of the model attended correctly (\Figref{fig:concept}b, top), this can be learned without additional bounding box supervision. At test time we evaluate the grounding performance, \ie whether the model assigned the highest weight to / attended to the correct bounding box.
The model is able to learn these associations as the parameters of the model are learned across all phrases and images. Thus, for a proper reconstruction, the visual semantic units and linguistic phrases have to match, \ie the models learns what certain visual phrases mean in the image. %

\input{grounding_examples_supervised}

\subsection{Semi-supervised and fully supervised grounding}
\label{sec:grounding:supervised}

\input{groundingresults}

If grounding supervision (phrase bounding box associations) is available, GroundeR \cite{rohrbach15arxiv1511.03745} can  integrate it by  adding a loss over the attention mechanism (\Figref{fig:concept}b, ``Attend''). Interestingly, this allows to provide supervision only for a subset of the phrases (semi-supervised) or all phrases (fully supervised).

For supervised grounding, \citet{plummer15iccv} 
 proposed to learn a CCA embedding \cite{gong2014eccv} between phrases and the visual representation. 
The Spatial Context Recurrent ConvNet (SCRC) \cite{hu16cvpr} and the approach of \citet{mao16cvpr} use a caption generation framework to score phrases on a set of bounding box proposals.
This allows to rank bounding box proposals for a given phrase or referential expression. 
\citet{hu16cvpr} show the benefit of transferring models trained on full-image description datasets as well as spatial (bounding box location and size) and full-image context features. \citet{mao16cvpr} show how to discriminatively train the caption generation framework to better distinguish different referential expression.

\subsection{Grounding results}
\label{sec:grounding:results}

In the following we discuss results  on the Flickr 30k Entities dataset \cite{plummer15iccv} and the ReferItGame dataset \cite{kazemzadeh14emnlp}, which both provide ground truth alignment between noun phrases (within sentences) and bounding boxes. 
For the unsupervised models, the grounding annotations are only used at test time for evaluation, not for training. %
All approaches use the activations of the second last layer of the VGG network \cite{simonyan15iclr} to encode the image inside the bounding boxes.

\Tableref{tbl:grounding}(a) compares the approaches quantitatively. The unsupervised variant of GroundeR reaches nearly the supervised performance of  CCA \cite{plummer15iccv} or SCRC\cite{hu16cvpr} on Flickr 30k Entities, successful examples are shown in \Figref{fig:flickr-unsupervised}. For the referential expressions of the ReferItGame dataset the unsupervised variant of GroundeR reaches performance on par with prior work (\Tableref{tbl:grounding}b) and quickly gains performance when adding few labeled training annotation (semi-supervised training).
In the fully supervised setting GroundeR improves significantly over state-of-the-art on both datasets, which is also reflected in the qualitative results shown in \Figref{fig:grounding:supervised}.

%% file: grounding_examples.tex
\begin{figure}[t]
\hspace{-1cm}
\begin{tabular}{@{}c@{\ }c@{\ }c@{}}
\includegraphics[clip=true,width=0.35\textwidth,height=0.2\textheight,keepaspectratio]{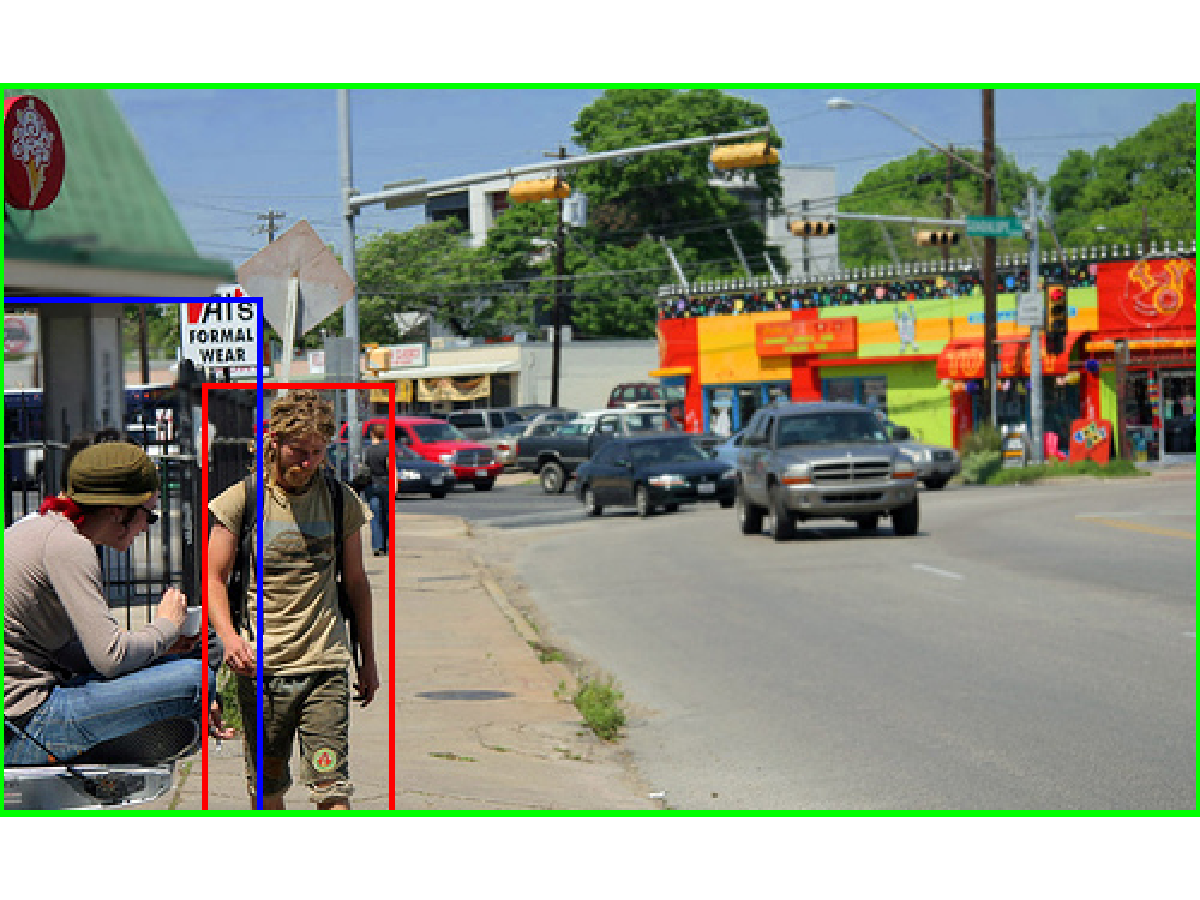} &
\includegraphics[clip=true,width=0.35\textwidth,height=0.2\textheight,keepaspectratio]{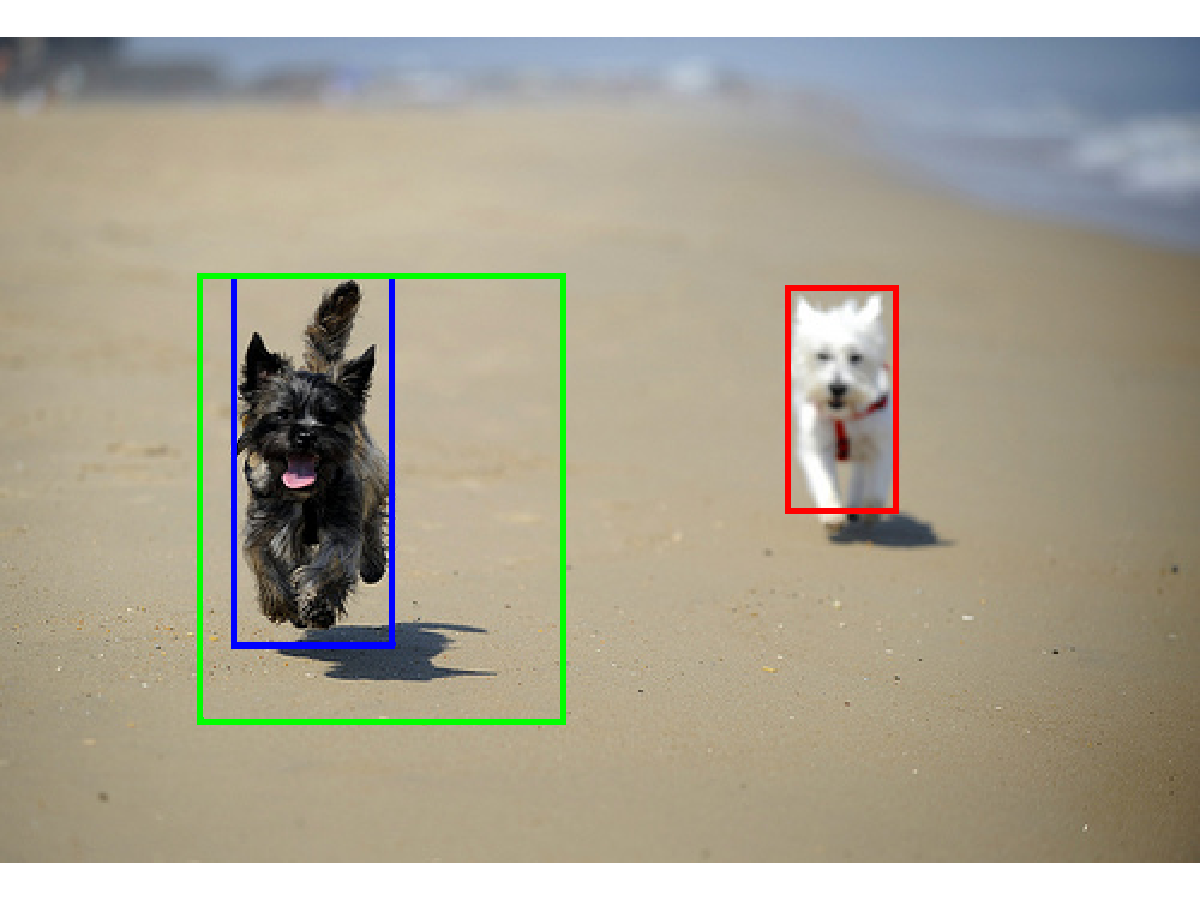} &
\includegraphics[clip=true,width=0.35\textwidth,height=0.2\textheight,keepaspectratio]{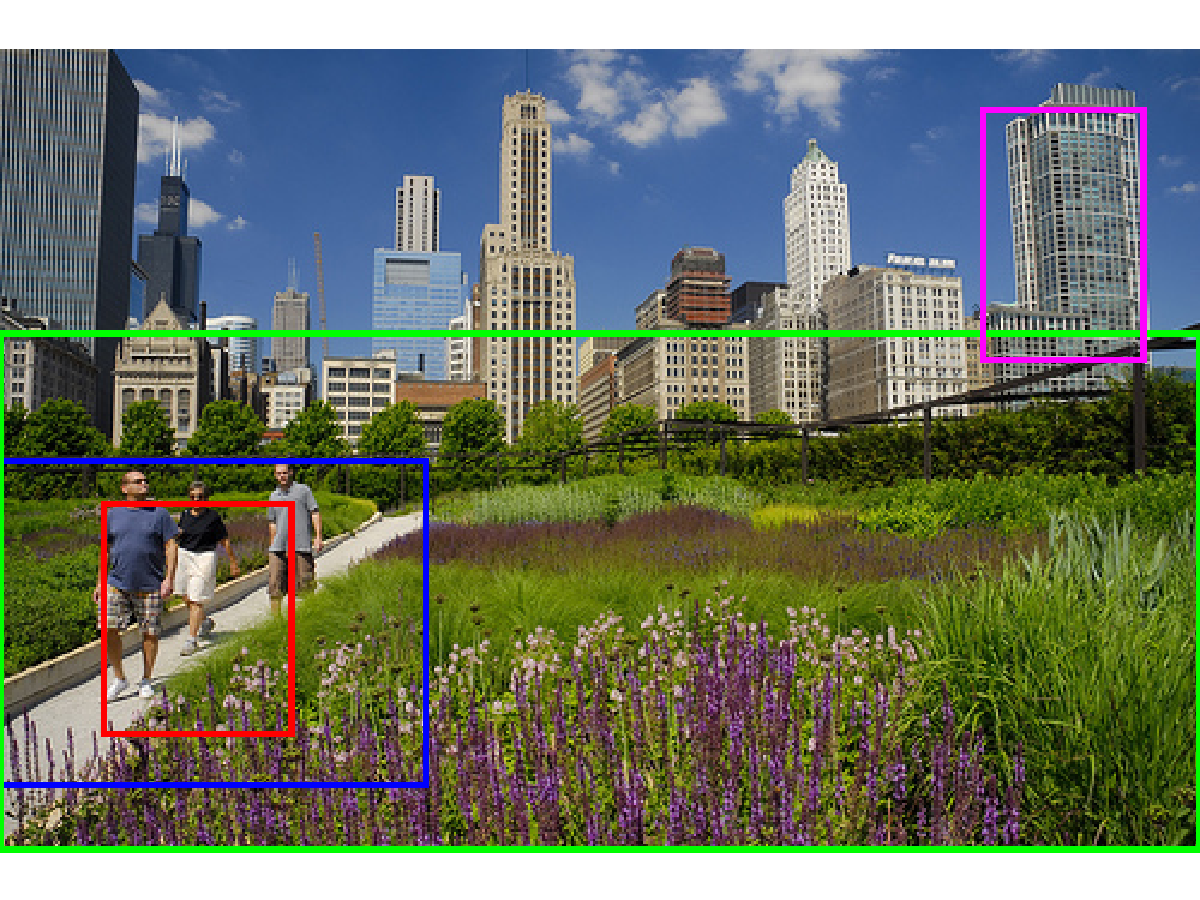} \\
\multicolumn{1}{m{4cm}}{\small{
\textcolor{red}{A man} walking by \textcolor{blue}{a sitting man} on \textcolor{green}{the street.}}} &
\multicolumn{1}{m{4cm}}{\small{
\textcolor{red}{A white dog} is following \textcolor{blue}{a black dog} along \textcolor{green}{the beach.}}} &
\multicolumn{1}{m{4cm}}{\small{
\textcolor{red}{Three people} on a walk down \textcolor{blue}{a cement path} beside \textcolor{green}{a field of wildflowers} with \textcolor{magenta}{skyscrapers} in the background.}} \vspace{4pt}\\
\end{tabular}
\caption{Qualitative results for GroundeR unsupervised \cite{rohrbach15arxiv1511.03745} on Flickr 30k Entities \cite{plummer15iccv}. Compact textual semantic units (phrases, \eg ``\emph{a sitting man}'') are associated with visual semantic units (bounding boxes). Best viewed in color.}
\label{fig:flickr-unsupervised}
\end{figure}

%% file: grounding_examples_supervised.tex
\begin{figure*}[t]
\hspace{-0.5cm}
\begin{tabular}{@{}c@{\ }c@{\ }c@{}}
\textbf{SCRC} \cite{hu16cvpr} & \textbf{GroundeR semi-supervised}  \cite{rohrbach15arxiv1511.03745}& \textbf{GroundeR supervised} \cite{rohrbach15arxiv1511.03745}\\
&with 3.12\% annot.&\\
\includegraphics[clip=true,width=0.35\textwidth,height=0.2\textheight,keepaspectratio]{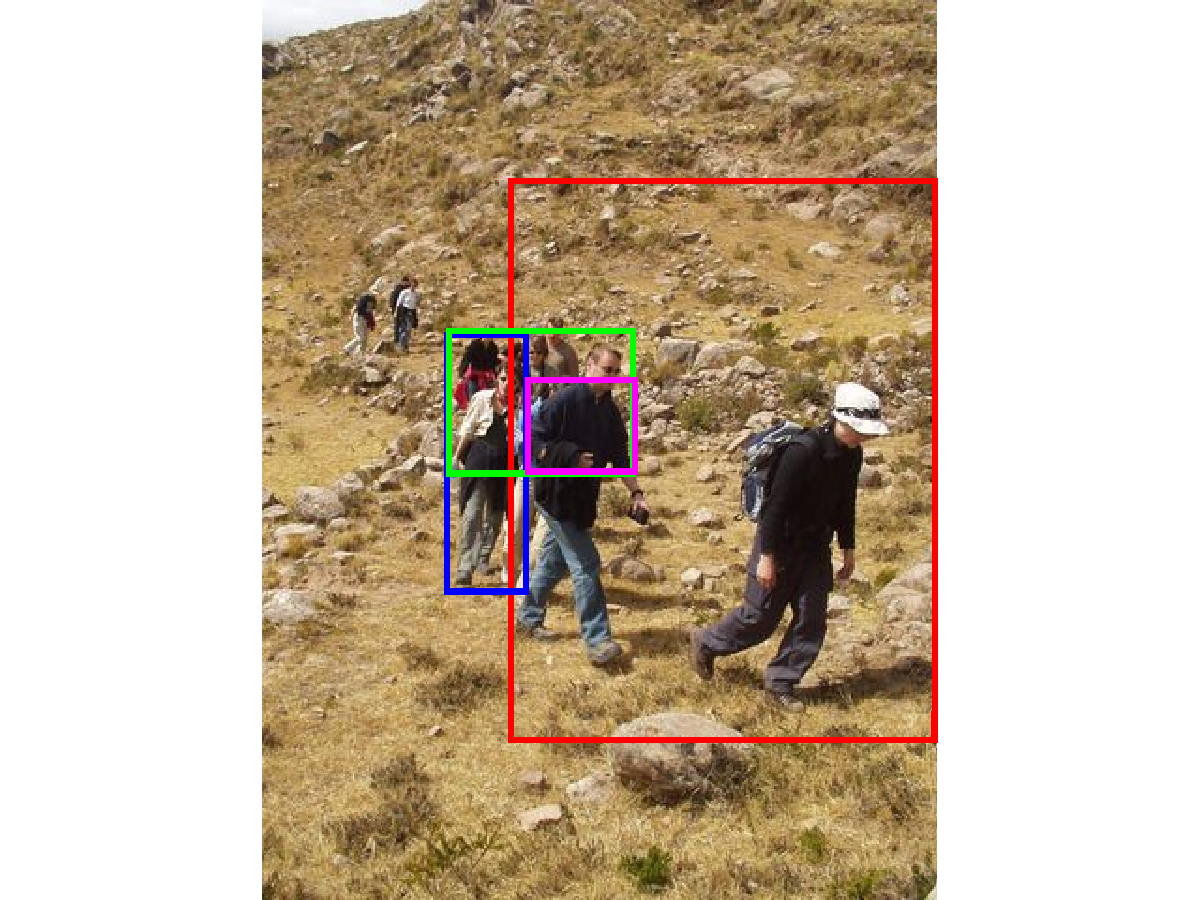} &
\includegraphics[clip=true,width=0.35\textwidth,height=0.2\textheight,keepaspectratio]{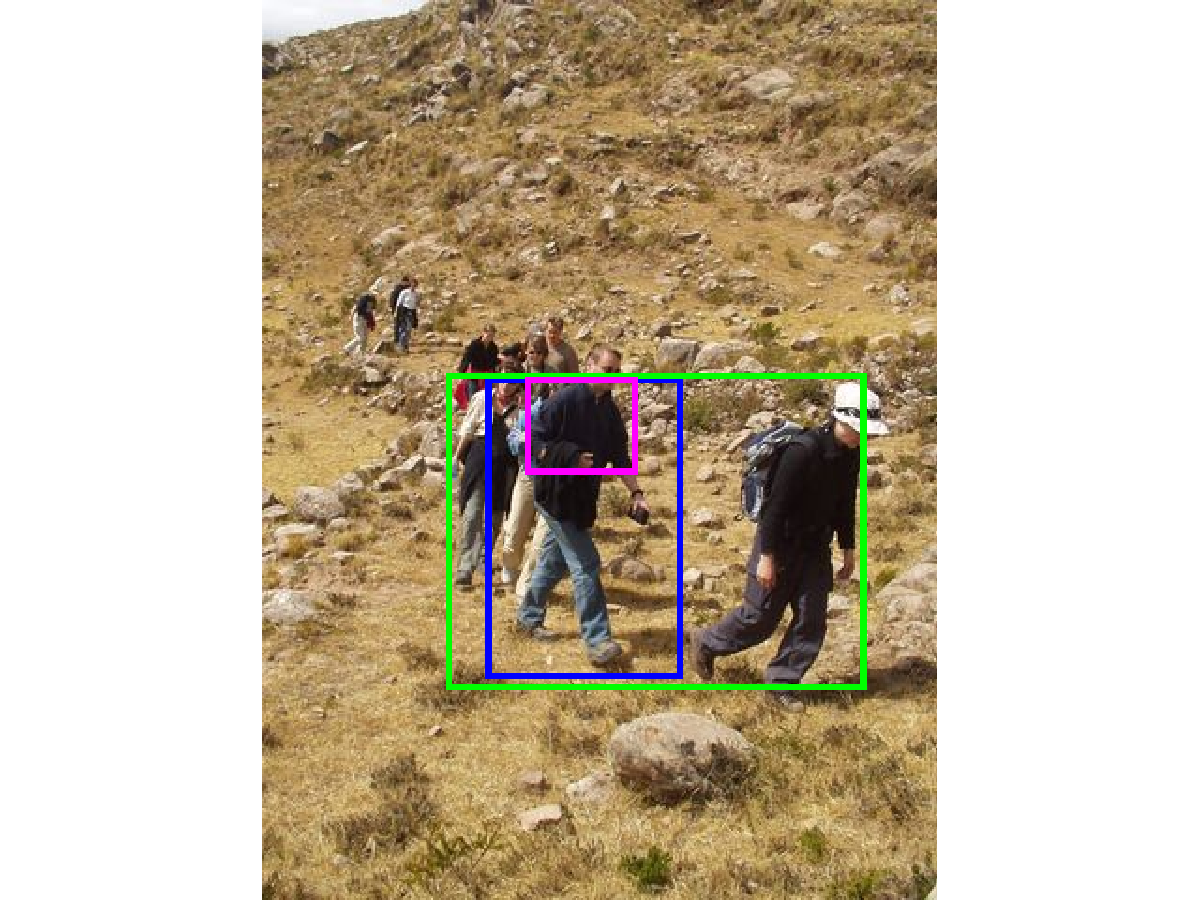} & 
\includegraphics[clip=true,width=0.35\textwidth,height=0.2\textheight,keepaspectratio]{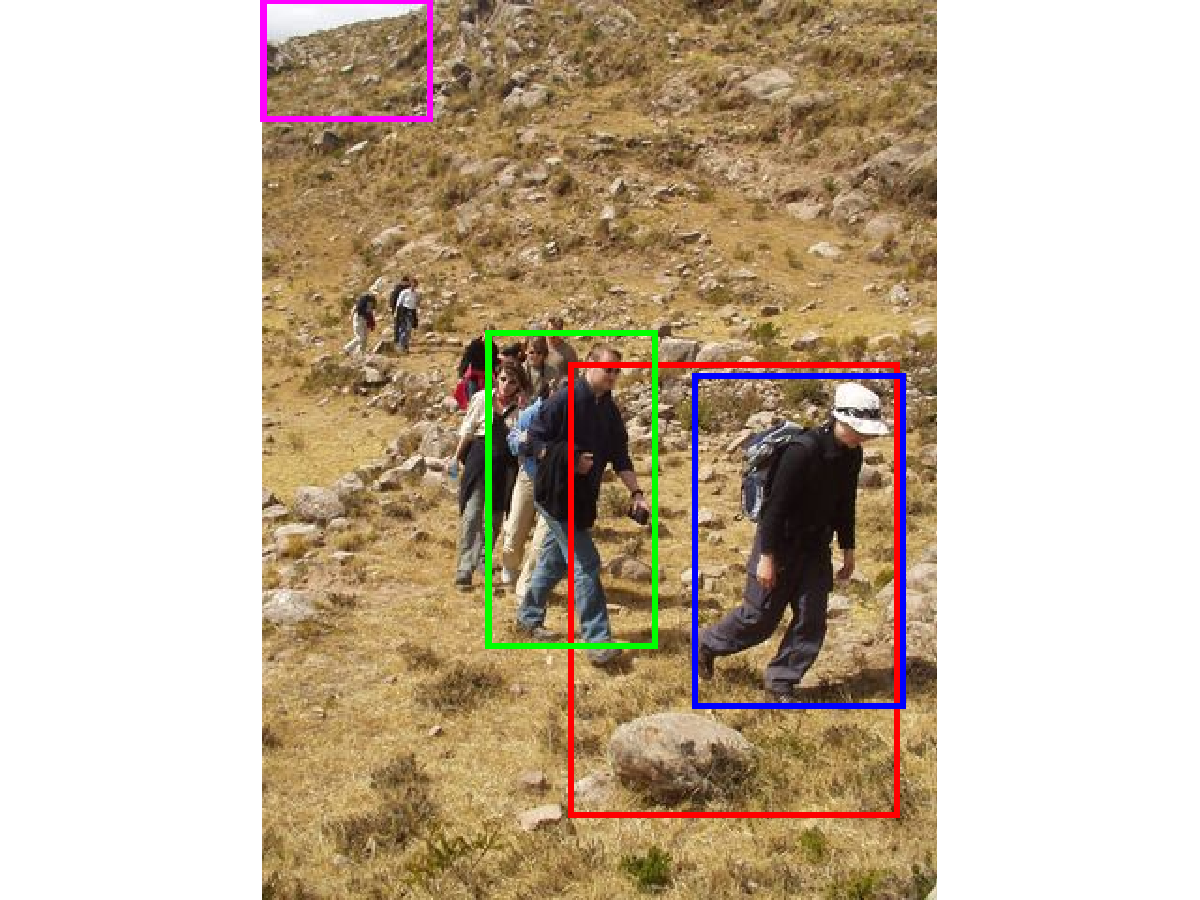} \\
\multicolumn{3}{m{12cm}}{\small{
\textcolor{red}{anywhere but the people} -- \textcolor{blue}{first person in line} -- \textcolor{green}{group people center}  -- \textcolor{magenta}{very top left of whole image}}}\\

\includegraphics[clip=true,width=0.35\textwidth,height=0.2\textheight,keepaspectratio]{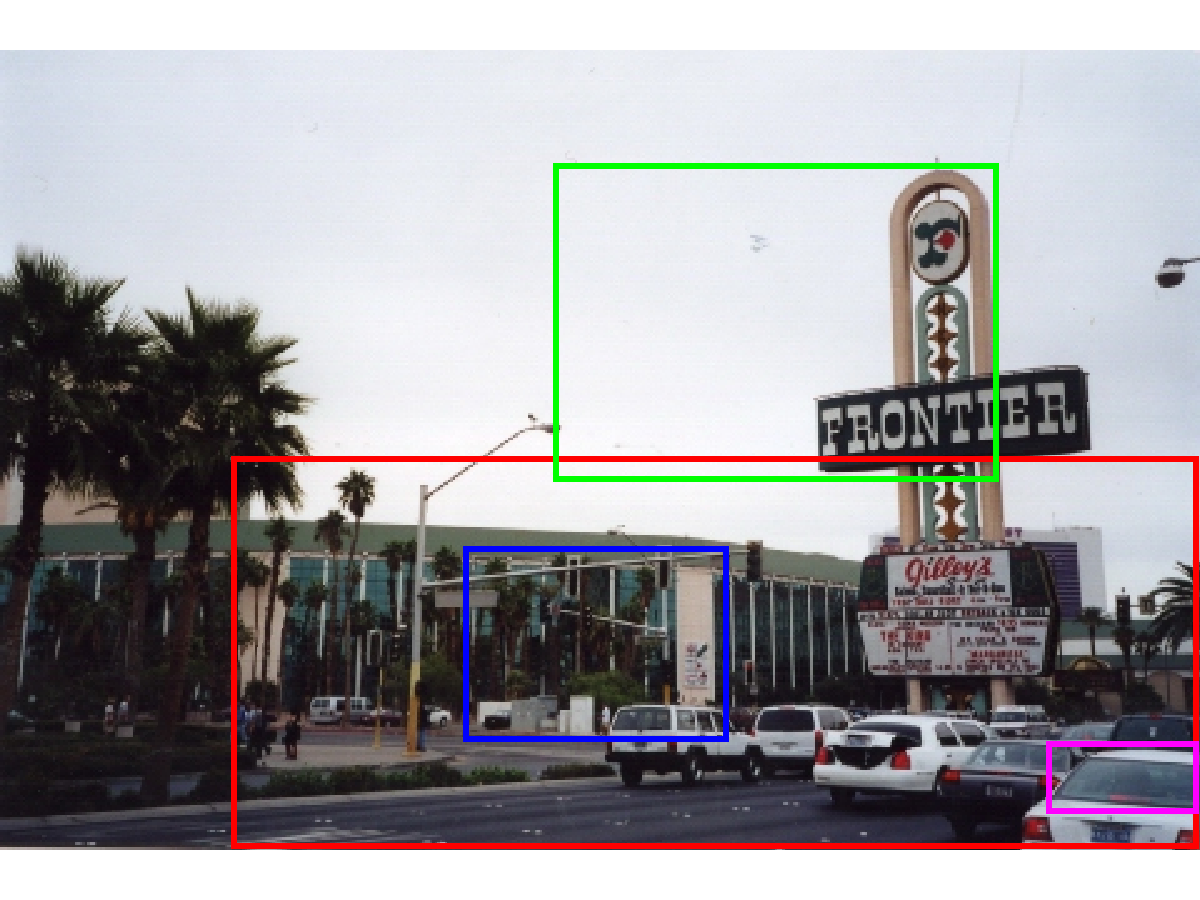} & 
\includegraphics[clip=true,width=0.35\textwidth,height=0.2\textheight,keepaspectratio]{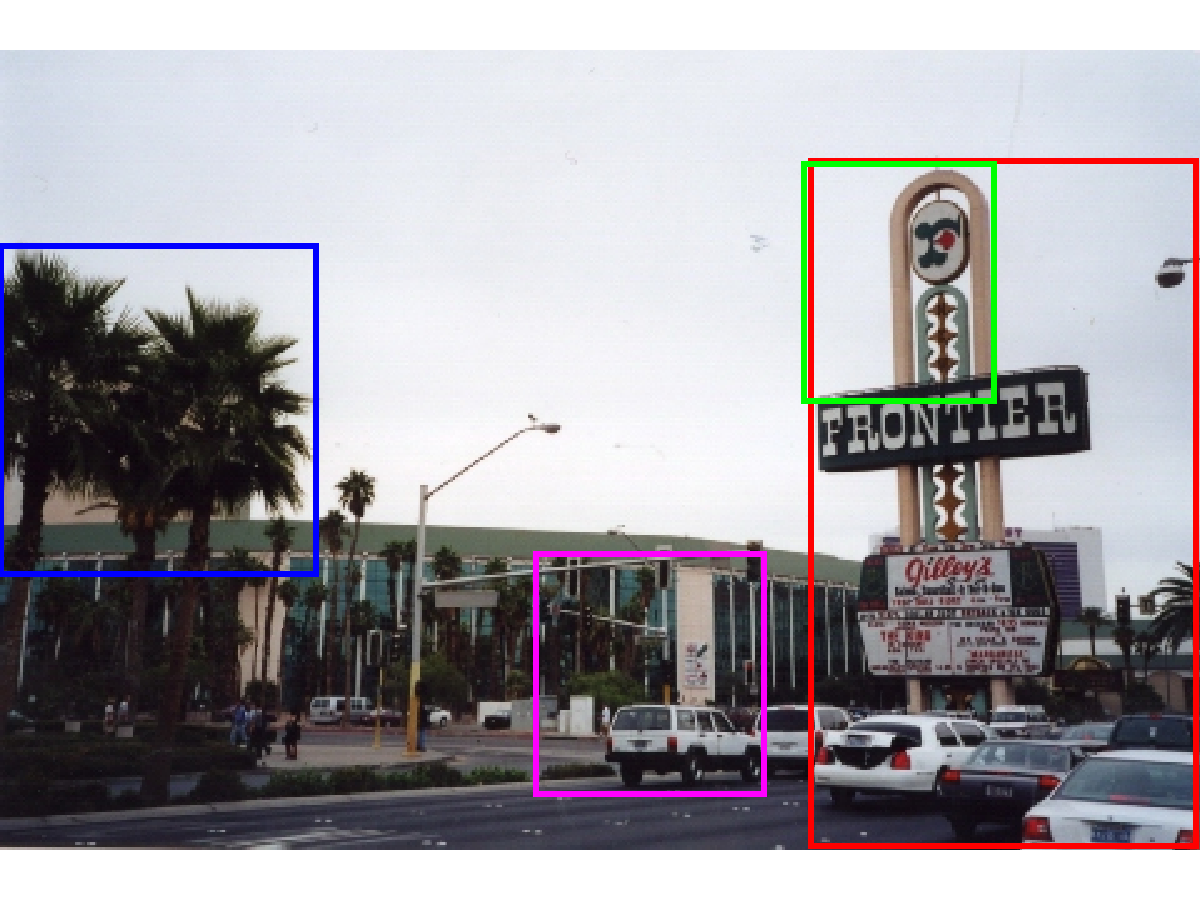} & 
\includegraphics[clip=true,width=0.35\textwidth,height=0.2\textheight,keepaspectratio]{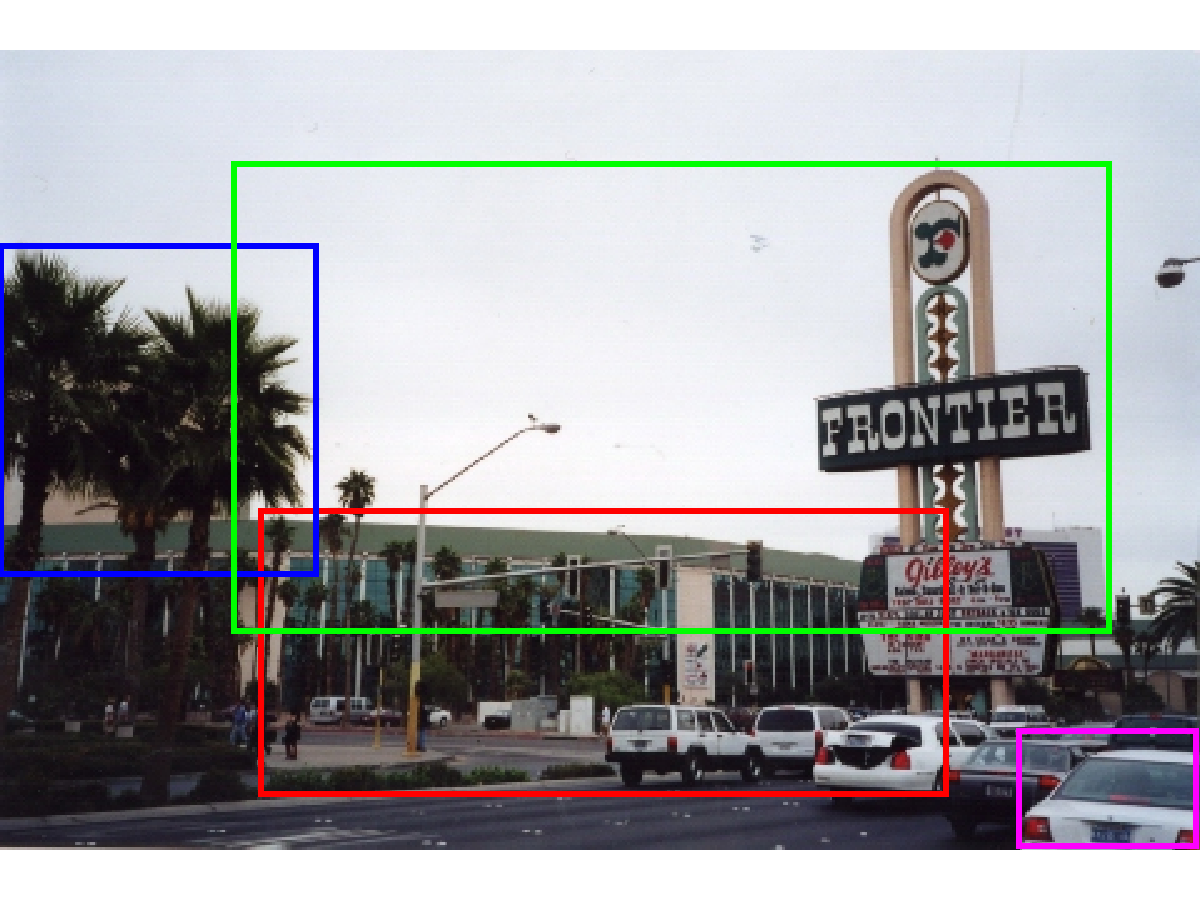} \vspace{-4pt}\\
\multicolumn{3}{m{12cm}}{\small{
\textcolor{red}{the street} -- \textcolor{blue}{tree to the far left} -- \textcolor{green}{top middle sky} -- \textcolor{magenta}{white car far right bottom corner}}} 
\end{tabular}
\caption{Qualitative grounding results on ReferItGame Dataset \cite{kazemzadeh14emnlp}. Different colors show different referential expressions for the same image. Best viewed in color.}
\label{fig:grounding:supervised}
\end{figure*}

%% file: groundingresults.tex
\newcommand{\midruleValLong}{\cmidrule(rr){1-1} \cmidrule(lr){2-3}}

\begin{table}[t]
\begin{tabular}{@{}p{0.45\textwidth}@{\quad}p{0.46\textwidth}@{}}
\input{test_table}
&
\input{grounding_test_table_referit}\\
\multicolumn{1}{c}{(a) Flickr 30k Entities dataset \cite{plummer15iccv}} & \multicolumn{1}{c}{(b) ReferItGame dataset \cite{kazemzadeh14emnlp}}\\
\end{tabular}
\caption{Phrase grounding, accuracy in \%.  VGG-CLS: Pre-training the VGG network \cite{simonyan15iclr} for the visual representation on ImageNet classification data only. VGG-DET: VGG further fine-tuned for the object detection task on  the PASCAL dataset \cite{everingham2010pascal} using Fast R-CNN \cite{girshick2015fast}. VGG+SPAT: VGG-CLS + spatial bounding box features (box location and size).}
\label{tbl:grounding}
\end{table}

%% file: test_table.tex
\newcommand{\midruleValShort}{\cmidrule(rr){1-1} \cmidrule(lr){2-2}}
\begin{tabular}{lc}
\toprule
Approach & Accuracy \\
\midruleValShort
\multicolumn{2}{l}{\textbf{Unsupervised training}} \\
GroundeR (VGG-CLS)  \cite{rohrbach15arxiv1511.03745}& 24.66 \\
GroundeR (VGG-DET)  \cite{rohrbach15arxiv1511.03745}& 32.42 \\
\midruleValShort
\multicolumn{2}{l}{\textbf{Semi-supervised training}} \\
GroundeR (VGG-CLS) \cite{rohrbach15arxiv1511.03745}\\
\ \ \ \ 3.12\% annotation  & 33.02 \\
\ \ \ \ 6.25\% annotation  & 37.10 \\
\ \ \ \ 12.5\% annotation  & 38.67 \\
\midruleValShort
\multicolumn{2}{l}{\textbf{Supervised training}} \\
CCA embedding \cite{plummer15iccv} & 25.30 \\
SCRC   (VGG+SPAT)  \cite{hu16cvpr}& 27.80 \\
GroundeR (VGG-CLS)  \cite{rohrbach15arxiv1511.03745}& 41.56 \\
GroundeR (VGG-DET) \cite{rohrbach15arxiv1511.03745}& 47.70 \\
\bottomrule\\
\end{tabular}

%% file: grounding_test_table_referit.tex
\newcommand{\midruleValShort}{\cmidrule(rr){1-1} \cmidrule(lr){2-2}}
\begin{tabular}{lc}
\toprule
Approach & Accuracy  \\
\midruleValShort
\multicolumn{2}{l}{\textbf{Unsupervised training}} \\
LRCN \cite{donahue15cvpr} (reported in \cite{hu16cvpr}) & 8.59  \\
CAFFE-7K \cite{guadarrama14rss} (reported in \cite{hu16cvpr}) & 10.38  \\
GroundeR (VGG+SPAT) \cite{rohrbach15arxiv1511.03745}  & 10.44  \\
\midruleValShort
\multicolumn{2}{l}{\textbf{Semi-supervised training}} \\
GroundeR (VGG+SPAT) \cite{rohrbach15arxiv1511.03745} \\
\ \ \ \ 3.12\% annotation & 15.03 \\
\ \ \ \ 6.25\% annotation & 19.53 \\
\ \ \ \ 12.5\% annotation & 21.65 \\
\midruleValShort
\multicolumn{2}{l}{\textbf{Supervised training}} \\
SCRC  (VGG+SPAT) \cite{hu16cvpr}& 17.93  \\
GroundeR (VGG+SPAT)  \cite{rohrbach15arxiv1511.03745} & 26.93 \\
\bottomrule\\\\
\end{tabular}

%% file: qa.tex
\section{Visual question answering}
\label{sec:qa}

\begin{figure}[t] \begin{center}
    \includegraphics[width=.6\linewidth]{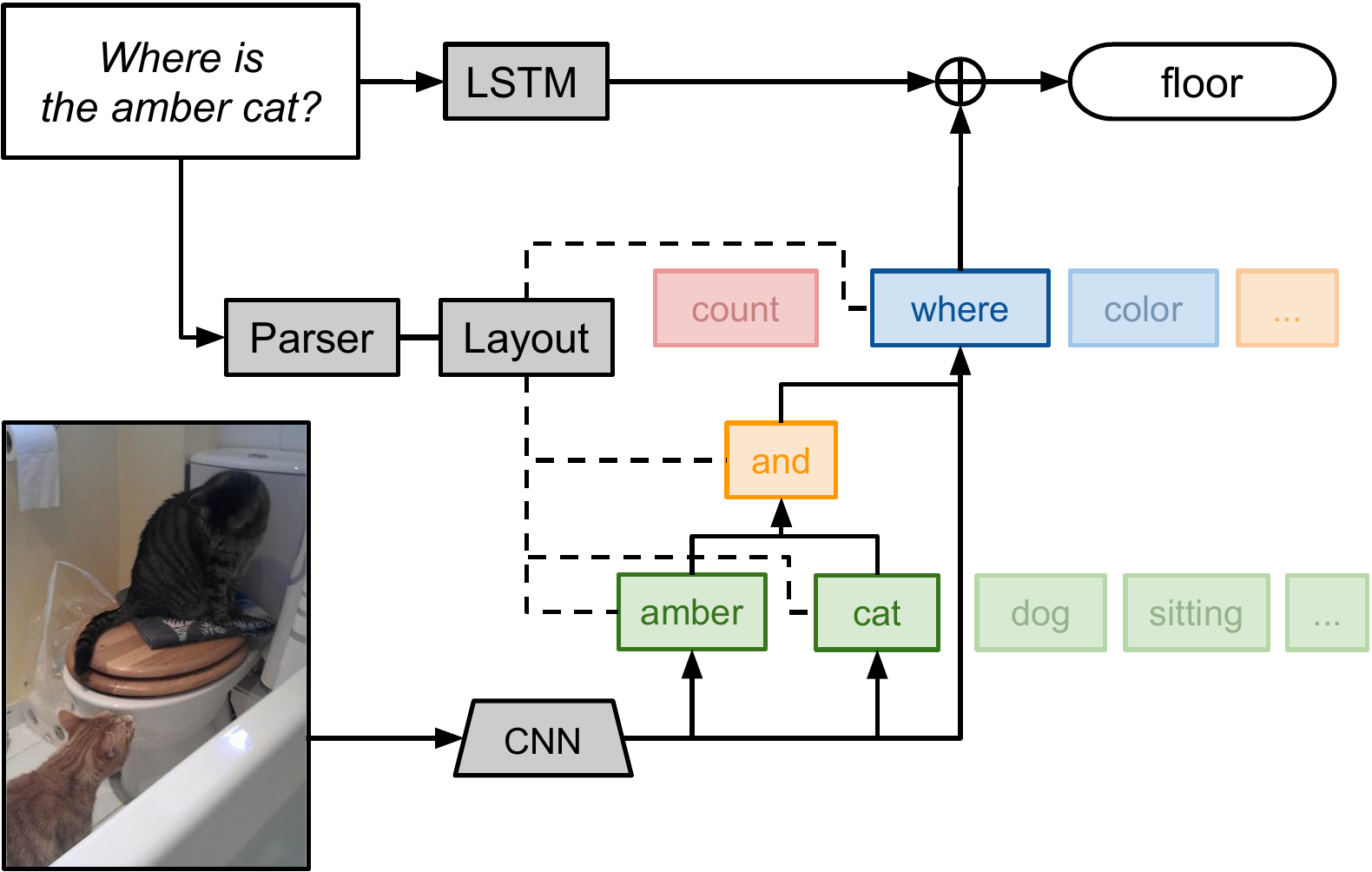} \end{center} \caption{To approach visual question answering, \citet{andreas16cvpr}  propose to dynamically create a deep network which is composed of different ``modules'' (colored boxes). These ``modules'' represent semantic units, \ie attributes, which link linguistic units in the question with computational units to do the corresponding visual recognition. 
     Adapted from \cite{andreas15arxiv}.  } \label{fig:qa:teaser}
\end{figure}

Visual question answering is the problem of answering natural language questions about images, \eg for the question \emph{``Where is the amber cat?''} about the image shown in \Figref{fig:qa:teaser} we want to predict the corresponding answer \emph{on the floor}, or just \emph{floor}. This is a very interesting problem with respect to several aspects. On the one hand it has many applications, such visual search, human-robot interaction, and assisting blind people. On the other hand, it is also an interesting research direction as it requires to relate textual and visual semantics. More specifically it requires to ground the question in the image, \eg by localizing the relevant part in the image (\emph{amber cat} in  \Figref{fig:qa:teaser}), and then recognizing and predicting an answer based on the question and the image content.
Consequently, this problem requires more complex semantic interaction between language and visual recognition than in previous sections, specifically, the problem requires ideas from grounding (\secref{sec:ground}) and recognition (\secref{sec:objectrecognition}) or description (\secref{sec:description}).%

\begin{figure}[t]
\renewcommand{\tabcolsep}{2pt}
  \footnotesize
  \hspace{-1cm}
  \begin{tabular}{p{0.51\textwidth}p{0.63\textwidth}}
   \begin{tabular}{lccccc}
    \toprule
    & \multicolumn{4}{c}{test-dev} & test \\
    \cmidrule(lr){2-5} \cmidrule(lr){6-6}
    & Y/N & Num & Other & All & All \\
    \midrule
    LSTM & 78.7 & 36.6 & 28.1 & 49.8 & -- \\
    ATT+LSTM & 80.6 & 36.4 & 42.0 & 57.2 & -- \\
    \midrule
    NMN & 70.7 & 36.8 & 39.2 & 54.8 & -- \\
    NMN+LSTM & 81.2 & 35.2 & 43.3 & 58.0 & -- \\
    NMN+LSTM+FT & 81.2 & 38.0 & 44.0 & 58.6 & \bf 58.7 \\
    \bottomrule
    \multicolumn{6}{l}{LSTM: a question-only baseline}\\
    \multicolumn{6}{l}{ATT: single \mod{find}+\mod{describe} for all questions}\\
    \multicolumn{6}{l}{NMN+LSTM: full model shown in
    \Figref{fig:qa:teaser}}\\ 
    \multicolumn{6}{l}{+FT: image features
    fine-tuned on  captions \cite{donahue15cvpr}}\\
    \multicolumn{6}{l}{NMN: ablation w/o LSTM}\\
    \bottomrule
  \end{tabular}&
  \renewcommand{\arraystretch}{1.4}
  \begin{tabular}{>{\raggedright}*{3}{p{.21\textwidth}}}
    \toprule
    \includegraphics[width=\linewidth]{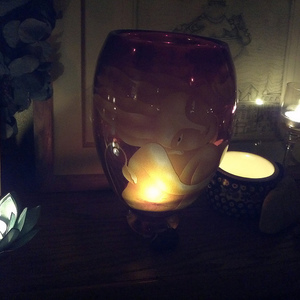} &
    \includegraphics[width=\linewidth]{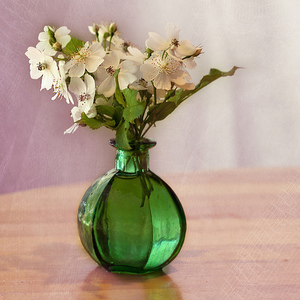} &
    \includegraphics[width=\linewidth]{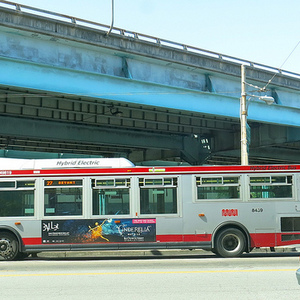} \\
    \emph{how many different lights in various different shapes and sizes?} \quad four (four)&
    \emph{what color is the vase?}  \newline  \newline green (green)&
    \emph{is the bus full of passengers?} \newline \newline   no (no)\\
    \bottomrule
  \end{tabular}\\
 (a) Results from evaluation server of \cite{antol15iccv} in \%. & (b) Answers from \cite{andreas15arxiv} (ground truth answers    in parentheses).\\
  \end{tabular}
  \caption{Results on the VQA dataset \cite{antol15iccv}. Adapted from \cite{andreas15arxiv}. }
    \label{fig:vqa-results}
      \label{fig:qa:qualresults}
\end{figure}

Most recent approaches to visual question answering learn a joint hidden  embedding of the question and the image to predict the answer  \cite{malinowski15iccv,ren15nips,gao15nips,antol15iccv} where all computation is shared and identical for all questions.
An exception to this is proposed by \citet{wu16arxiv1603.02814}, who learn an intermediate attribute representation from the image descriptions, similar to the work discussed in \secsref{sec:moviedescription} and \ref{sec:novel_sentences}. Interestingly, this intermediate layer of attributes allows to query an external knowledge base to provide additional (textual) information not visible in the image. The embedded textual knowledge base information is combined with the attribute representation and the hidden representation of a caption-generation recurrent network (LSTM) and forms the input to an LSTM-based question-answer encoder-decoder \cite{malinowski15iccv}.

\citet{andreas16cvpr} go one step further with respect to compositionality and propose to predict a compositional neural network structure from the questions. As visualized in \Figref{fig:qa:teaser}  the question \emph{``Where is the amber cat?''} is decomposed into network ``modules'' \mod{amber}, \mod{cat}, \mod{and}, and \mod{where}. These modules are semantic units, \ie attributes, which connect most relevant semantic components of the questions (\ie word or short phrases) with corresponding computation to recognize it in the image.
These \nmn{} (NMN) have different types of modules for different types of attributes. Different types have different colors in \Figref{fig:qa:teaser}. %
The \mod{find[$cat$]} and \mod{find[$amber$]} (green) modules take in CNN activations (VGG \cite{simonyan15iclr}, last convolutional layer) and produce a spatial attention heatmap, while  \mod{combine[$and$]} (orange) combines two heatmaps to a single one, and \mod{describe[$where$]} (blue) takes in a heatmap and CNN features to predict an answer.  Note that the distinction between different types, \eg\ \mod{find} versus \mod{describe}, which have different kind of computation and different instances, \eg\ \mod{find[$cat$]} versus \mod{find[$amber$]}, which learn different parameters. All parameters are initialized randomly and only trained from question answer pairs.
Interestingly, in this work attributes are not only distinguished with respect of their type, but also are composed with other attributes in a deep network, whose parameters' are learned end-to-end from examples, here question-answer pairs. 
In a follow up work, \citet{andreas16naacl} learn not only the modules, but also  what the best  network structure is from a set of parser proposals, using reinforcement learning.

In addition to NMN, \citeauthor{andreas16cvpr} \cite{andreas16cvpr,andreas16naacl} also incorporate a recurrent network (LSTM) to model common sense knowledge and dataset bias which has been shown to be important for visual question answering \cite{malinowski15iccv}. Quantitative results in \Tableref{fig:vqa-results}(a) indicate that NMNs are indeed a powerful tool to question answering, a few qualitative results can be seen \figref{fig:qa:qualresults}(b).

%% file: rohrbach16chapter.bbl
\begin{thebibliography}{102}
\providecommand{\natexlab}[1]{#1}
\providecommand{\url}[1]{\texttt{#1}}
\expandafter\ifx\csname urlstyle\endcsname\relax
  \providecommand{\doi}[1]{doi: #1}\else
  \providecommand{\doi}{doi: \begingroup \urlstyle{rm}\Url}\fi

\bibitem[Andreas et~al.(2015)Andreas, Rohrbach, Darrell, and
  Klein]{andreas15arxiv}
J.~Andreas, M.~Rohrbach, T.~Darrell, and D.~Klein.
\newblock Deep compositional question answering with neural module networks.
\newblock \emph{arXiv preprint arXiv:1511.02799}, 2015.

\bibitem[Andreas et~al.(2016{\natexlab{a}})Andreas, Rohrbach, Darrell, and
  Klein]{andreas16cvpr}
J.~Andreas, M.~Rohrbach, T.~Darrell, and D.~Klein.
\newblock Neural module networks.
\newblock In \emph{Conference on Computer Vision and Pattern Recognition
  (CVPR)}, 2016{\natexlab{a}}.

\bibitem[Andreas et~al.(2016{\natexlab{b}})Andreas, Rohrbach, Darrell, and
  Klein]{andreas16naacl}
J.~Andreas, M.~Rohrbach, T.~Darrell, and D.~Klein.
\newblock Learning to compose neural networks for question answering.
\newblock In \emph{Proceedings of the Conference of the North American Chapter
  of the Association for Computational Linguistics (NAACL)},
  2016{\natexlab{b}}.

\bibitem[Antol et~al.(2015)Antol, Agrawal, Lu, Mitchell, Batra, Zitnick, and
  Parikh]{antol15iccv}
S.~Antol, A.~Agrawal, J.~Lu, M.~Mitchell, D.~Batra, C.~L. Zitnick, and
  D.~Parikh.
\newblock Vqa: Visual question answering.
\newblock In \emph{International Conference on Computer Vision (ICCV)}, 2015.

\bibitem[Barnard et~al.(2003)Barnard, Duygulu, Forsyth, De~Freitas, Blei, and
  Jordan]{barnard03jmlr}
K.~Barnard, P.~Duygulu, D.~Forsyth, N.~De~Freitas, D.~M. Blei, and M.~I.
  Jordan.
\newblock Matching words and pictures.
\newblock \emph{Journal of Machine Learning Research (JMLR)}, 3:\penalty0
  1107--1135, 2003.

\bibitem[Bart and Ullman(2005)]{bart05bmvc}
E.~Bart and S.~Ullman.
\newblock Single-example learning of novel classes using representation by
  similarity.
\newblock In \emph{Proceedings of the British Machine Vision Conference
  (BMVC)}, 2005.

\bibitem[Chen et~al.(2006)Chen, Lin, and Wei]{chen06acl}
H.-H. Chen, M.-S. Lin, and Y.-C. Wei.
\newblock Novel association measures using web search with double checking.
\newblock In \emph{Proceedings of the Annual Meeting of the Association for
  Computational Linguistics (ACL)}, 2006.

\bibitem[Chen et~al.(2015)Chen, Fang, Lin, Vedantam, Gupta, Dollar, and
  Zitnick]{chen15arXiv1504.00325}
X.~Chen, H.~Fang, T.-Y. Lin, R.~Vedantam, S.~Gupta, P.~Dollar, and C.~L.
  Zitnick.
\newblock Microsoft {COCO} captions: Data collection and evaluation server.
\newblock \emph{arXiv preprint arXiv:1504.00325}, 2015.

\bibitem[Deng et~al.(2009)Deng, Dong, Socher, Li, Li, and Fei-Fei]{deng09cvpr}
J.~Deng, W.~Dong, R.~Socher, L.-J. Li, K.~Li, and L.~Fei-Fei.
\newblock Imagenet: A large-scale hierarchical image database.
\newblock In \emph{Conference on Computer Vision and Pattern Recognition
  (CVPR)}, 2009.

\bibitem[Deng et~al.(2010)Deng, Berg, Li, and Fei-Fei]{deng10eccv}
J.~Deng, A.~Berg, K.~Li, and L.~Fei-Fei.
\newblock What does classifying more than 10,000 image categories tell us?
\newblock In \emph{European Conference on Computer Vision (ECCV)}, 2010.

\bibitem[Dice(1945)]{dice45ecology}
L.~R. Dice.
\newblock Measures of the amount of ecologic association between species.
\newblock \emph{Ecology}, 26\penalty0 (3):\penalty0 297--302, 1945.

\bibitem[Donahue et~al.(2015)Donahue, Hendricks, Guadarrama, Rohrbach,
  Venugopalan, Saenko, and Darrell]{donahue15cvpr}
J.~Donahue, L.~A. Hendricks, S.~Guadarrama, M.~Rohrbach, S.~Venugopalan,
  K.~Saenko, and T.~Darrell.
\newblock Long-term recurrent convolutional networks for visual recognition and
  description.
\newblock In \emph{Conference on Computer Vision and Pattern Recognition
  (CVPR)}, 2015.

\bibitem[Duan et~al.(2012)Duan, Parikh, Crandall, and Grauman]{duan12cvpr}
K.~Duan, D.~Parikh, D.~Crandall, and K.~Grauman.
\newblock {Discovering Localized Attributes for Fine-grained Recognition}.
\newblock In \emph{Conference on Computer Vision and Pattern Recognition
  (CVPR)}, 2012.

\bibitem[Ebert et~al.(2010)Ebert, Larlus, and Schiele]{ebert10eccv}
S.~Ebert, D.~Larlus, and B.~Schiele.
\newblock {Extracting Structures in Image Collections for Object Recognition}.
\newblock In \emph{European Conference on Computer Vision (ECCV)}, 2010.

\bibitem[Everingham et~al.(2010)Everingham, Van~Gool, Williams, Winn, and
  Zisserman]{everingham2010pascal}
M.~Everingham, L.~Van~Gool, C.~K. Williams, J.~Winn, and A.~Zisserman.
\newblock The pascal visual object classes (voc) challenge.
\newblock \emph{International Journal of Computer Vision (IJCV)}, 88\penalty0
  (2):\penalty0 303--338, 2010.

\bibitem[Fang et~al.(2015)Fang, Gupta, Iandola, Srivastava, Deng, Doll{\'{a}}r,
  Gao, He, Mitchell, Platt, Zitnick, and Zweig]{fang15cvpr}
H.~Fang, S.~Gupta, F.~N. Iandola, R.~Srivastava, L.~Deng, P.~Doll{\'{a}}r,
  J.~Gao, X.~He, M.~Mitchell, J.~C. Platt, C.~L. Zitnick, and G.~Zweig.
\newblock From captions to visual concepts and back.
\newblock In \emph{Conference on Computer Vision and Pattern Recognition
  (CVPR)}, 2015.

\bibitem[Farhadi et~al.(2009)Farhadi, Endres, Hoiem, and Forsyth]{Farhadi2009}
A.~Farhadi, I.~Endres, D.~Hoiem, and D.~Forsyth.
\newblock {Describing objects by their attributes}.
\newblock In \emph{Conference on Computer Vision and Pattern Recognition
  (CVPR)}, 2009.

\bibitem[Farhadi et~al.(2010)Farhadi, Endres, and Hoiem]{farhadi10cvpr}
A.~Farhadi, I.~Endres, and D.~Hoiem.
\newblock Attribute-centric recognition for cross-category generalization.
\newblock In \emph{Conference on Computer Vision and Pattern Recognition
  (CVPR)}, 2010.

\bibitem[Farrell et~al.(2011)Farrell, Oza, Morariu, Darrell, and
  Davis]{Farrell2011}
R.~Farrell, O.~Oza, V.~Morariu, T.~Darrell, and L.~Davis.
\newblock Birdlets: Subordinate categorization using volumetric primitives and
  pose-normalized appearance.
\newblock In \emph{International Conference on Computer Vision (ICCV)}, 2011.

\bibitem[Fellbaum(1998)]{fellbaum:wordnet}
C.~Fellbaum.
\newblock \emph{WordNet: An Electronical Lexical Database}.
\newblock The MIT Press, 1998.

\bibitem[Frome et~al.(2013)Frome, Corrado, Shlens, Bengio, Dean, Ranzato, and
  Mikolov]{frome13nips}
A.~Frome, G.~Corrado, J.~Shlens, S.~Bengio, J.~Dean, M.~Ranzato, and
  T.~Mikolov.
\newblock Devise: A deep visual-semantic embedding model.
\newblock In \emph{Conference on Neural Information Processing Systems (NIPS)},
  2013.

\bibitem[Fu et~al.(2014)Fu, Hospedales, Xiang, and Gong]{fu13pami}
Y.~Fu, T.~M. Hospedales, T.~Xiang, and S.~Gong.
\newblock Learning multimodal latent attributes.
\newblock \emph{IEEE Transactions on Pattern Analysis and Machine Intelligence
  (PAMI)}, 36\penalty0 (2):\penalty0 303--316, 2014.

\bibitem[Gabrilovich and Markovitch(2007)]{gabrilovich07ijcai}
E.~Gabrilovich and S.~Markovitch.
\newblock {Computing Semantic Relatedness using Wikipedia-based Explicit
  Semantic Analysis}.
\newblock In \emph{Proceedings of the International Joint Conference on
  Artificial Intelligence (IJCAI)}, 2007.

\bibitem[Gao et~al.(2015)Gao, Mao, Zhou, Huang, Wang, and Xu]{gao15nips}
H.~Gao, J.~Mao, J.~Zhou, Z.~Huang, L.~Wang, and W.~Xu.
\newblock Are you talking to a machine? dataset and methods for multilingual
  image question answering.
\newblock In \emph{Conference on Neural Information Processing Systems (NIPS)},
  2015.

\bibitem[Girshick(2015)]{girshick2015fast}
R.~Girshick.
\newblock Fast {R-CNN}.
\newblock In \emph{International Conference on Computer Vision (ICCV)}, 2015.

\bibitem[Gong et~al.(2014)Gong, Wang, Hodosh, Hockenmaier, and
  Lazebnik]{gong2014eccv}
Y.~Gong, L.~Wang, M.~Hodosh, J.~Hockenmaier, and S.~Lazebnik.
\newblock Improving image-sentence embeddings using large weakly annotated
  photo collections.
\newblock In \emph{European Conference on Computer Vision (ECCV)}, 2014.

\bibitem[Guadarrama et~al.(2014)Guadarrama, Rodner, Saenko, Zhang, Farrell,
  Donahue, and Darrell]{guadarrama14rss}
S.~Guadarrama, E.~Rodner, K.~Saenko, N.~Zhang, R.~Farrell, J.~Donahue, and
  T.~Darrell.
\newblock Open-vocabulary object retrieval.
\newblock In \emph{Robotics: science and systems}, 2014.

\bibitem[He et~al.(2015)He, Zhang, Ren, and Sun]{he15iccv}
K.~He, X.~Zhang, S.~Ren, and J.~Sun.
\newblock Delving deep into rectifiers: Surpassing human-level performance on
  imagenet classification.
\newblock In \emph{International Conference on Computer Vision (ICCV)}, 2015.

\bibitem[Hendricks et~al.(2015)Hendricks, Venugopalan, Rohrbach, Mooney,
  Saenko, and Darrell]{hendricks15arxiv}
L.~A. Hendricks, S.~Venugopalan, M.~Rohrbach, R.~Mooney, K.~Saenko, and
  T.~Darrell.
\newblock Deep compositional captioning: Describing novel object categories
  without paired training data.
\newblock \emph{arXiv preprint arXiv:1511.05284v1}, 2015.

\bibitem[Hendricks et~al.(2016)Hendricks, Venugopalan, Rohrbach, Mooney,
  Saenko, and Darrell]{hendricks16cvpr}
L.~A. Hendricks, S.~Venugopalan, M.~Rohrbach, R.~Mooney, K.~Saenko, and
  T.~Darrell.
\newblock Deep compositional captioning: Describing novel object categories
  without paired training data.
\newblock In \emph{Conference on Computer Vision and Pattern Recognition
  (CVPR)}, 2016.

\bibitem[Hoffman et~al.(2014)Hoffman, Guadarrama, Tzeng, Donahue, Girshick,
  Darrell, and Saenko]{hoffman14nips}
J.~Hoffman, S.~Guadarrama, E.~Tzeng, J.~Donahue, R.~Girshick, T.~Darrell, and
  K.~Saenko.
\newblock {LSDA}: Large scale detection through adaptation.
\newblock In \emph{Conference on Neural Information Processing Systems (NIPS)},
  2014.

\bibitem[Hu et~al.(2015)Hu, Xu, Rohrbach, Feng, Saenko, and Darrell]{hu16cvpr}
R.~Hu, H.~Xu, M.~Rohrbach, J.~Feng, K.~Saenko, and T.~Darrell.
\newblock Natural language object retrieval.
\newblock In \emph{Conference on Computer Vision and Pattern Recognition
  (CVPR)}, 2015.

\bibitem[Hu et~al.(2016)Hu, Rohrbach, and Darrell]{hu16arxiv1603.06180}
R.~Hu, M.~Rohrbach, and T.~Darrell.
\newblock Segmentation from natural language expressions.
\newblock \emph{arXiv preprint arXiv:1603.06180}, 2016.

\bibitem[Johnson et~al.(2015)Johnson, Krishna, Stark, Li, Shamma, Bernstein,
  and Fei-Fei]{johnson2015cvpr}
J.~Johnson, R.~Krishna, M.~Stark, L.-J. Li, D.~Shamma, M.~Bernstein, and
  L.~Fei-Fei.
\newblock Image retrieval using scene graphs.
\newblock In \emph{Conference on Computer Vision and Pattern Recognition
  (CVPR)}, 2015.

\bibitem[Johnson et~al.(2016)Johnson, Karpathy, and Fei-Fei]{johnson16cvpr}
J.~Johnson, A.~Karpathy, and L.~Fei-Fei.
\newblock Densecap: Fully convolutional localization networks for dense
  captioning.
\newblock In \emph{Conference on Computer Vision and Pattern Recognition
  (CVPR)}, 2016.

\bibitem[Karpathy and Fei-Fei(2015)]{karpathy15cvpr}
A.~Karpathy and L.~Fei-Fei.
\newblock Deep visual-semantic alignments for generating image descriptions.
\newblock In \emph{Conference on Computer Vision and Pattern Recognition
  (CVPR)}, 2015.

\bibitem[Karpathy et~al.(2014)Karpathy, Joulin, and Fei-Fei]{karpathy14nips}
A.~Karpathy, A.~Joulin, and L.~Fei-Fei.
\newblock Deep fragment embeddings for bidirectional image sentence mapping.
\newblock In \emph{Conference on Neural Information Processing Systems (NIPS)},
  2014.

\bibitem[Kazemzadeh et~al.(2014)Kazemzadeh, Ordonez, Matten, and
  Berg]{kazemzadeh14emnlp}
S.~Kazemzadeh, V.~Ordonez, M.~Matten, and T.~L. Berg.
\newblock Referitgame: Referring to objects in photographs of natural scenes.
\newblock In \emph{Proceedings of the Conference on Empirical Methods in
  Natural Language Processing (EMNLP)}, 2014.

\bibitem[Koehn(2010)]{koehn10book}
P.~Koehn.
\newblock \emph{Statistical Machine Translation}.
\newblock Cambridge University Press, 2010.

\bibitem[Kong et~al.(2014)Kong, Lin, Bansal, Urtasun, and Fidler]{kong14cvpr}
C.~Kong, D.~Lin, M.~Bansal, R.~Urtasun, and S.~Fidler.
\newblock What are you talking about? text-to-image coreference.
\newblock In \emph{Conference on Computer Vision and Pattern Recognition
  (CVPR)}, 2014.

\bibitem[Krishna et~al.(2016)Krishna, Zhu, Groth, Johnson, Hata, Kravitz, Chen,
  Kalanditis, Li, Shamma, Bernstein, and Fei-Fei]{krishnavisualgenome}
R.~Krishna, Y.~Zhu, O.~Groth, J.~Johnson, K.~Hata, J.~Kravitz, S.~Chen,
  Y.~Kalanditis, L.-J. Li, D.~A. Shamma, M.~Bernstein, and L.~Fei-Fei.
\newblock Visual genome: Connecting language and vision using crowdsourced
  dense image annotations.
\newblock \emph{arXiv preprint arXiv:1602.07332}, 2016.

\bibitem[Krizhevsky et~al.(2012)Krizhevsky, Sutskever, and
  Hinton]{krizhevsky12nips}
A.~Krizhevsky, I.~Sutskever, and G.~E. Hinton.
\newblock Imagenet classification with deep convolutional neural networks.
\newblock In \emph{Conference on Neural Information Processing Systems (NIPS)},
  2012.

\bibitem[Lampert et~al.(2009)Lampert, Nickisch, and Harmeling]{lampert09cvpr}
C.~Lampert, H.~Nickisch, and S.~Harmeling.
\newblock {Learning to detect unseen object classes by between-class attribute
  transfer}.
\newblock In \emph{Conference on Computer Vision and Pattern Recognition
  (CVPR)}, 2009.

\bibitem[Lampert et~al.(2014)Lampert, Nickisch, and Harmeling]{lampert13pami}
C.~H. Lampert, H.~Nickisch, and S.~Harmeling.
\newblock Attribute-based classification for zero-shot visual object
  categorization.
\newblock \emph{IEEE Transactions on Pattern Analysis and Machine Intelligence
  (PAMI)}, 36\penalty0 (3):\penalty0 453--465, 2014.

\bibitem[Liang et~al.(2013)Liang, Xu, Cheng, Min, and Lu]{liang13tmm}
C.~Liang, C.~Xu, J.~Cheng, W.~Min, and H.~Lu.
\newblock Script-to-movie: A computational framework for story movie
  composition.
\newblock \emph{Multimedia, IEEE Transactions on}, 15\penalty0 (2):\penalty0
  401--414, 2013.

\bibitem[Lin(1998)]{lin98icml}
D.~Lin.
\newblock An information-theoretic definition of similarity.
\newblock In \emph{International Conference on Machine Learning (ICML)}, 1998.

\bibitem[Lin et~al.(2014)Lin, Maire, Belongie, Hays, Perona, Ramanan,
  Doll{\'a}r, and Zitnick]{coco2014}
T.-Y. Lin, M.~Maire, S.~Belongie, J.~Hays, P.~Perona, D.~Ramanan,
  P.~Doll{\'a}r, and C.~L. Zitnick.
\newblock Microsoft coco: Common objects in context.
\newblock In \emph{European Conference on Computer Vision (ECCV)}, 2014.

\bibitem[Malinowski and Fritz(2014)]{malinowski14nips}
M.~Malinowski and M.~Fritz.
\newblock A multi-world approach to question answering about real-world scenes
  based on uncertain input.
\newblock In \emph{Conference on Neural Information Processing Systems (NIPS)},
  2014.

\bibitem[Malinowski et~al.(2015)Malinowski, Rohrbach, and
  Fritz]{malinowski15iccv}
M.~Malinowski, M.~Rohrbach, and M.~Fritz.
\newblock Ask your neurons: A neural-based approach to answering questions
  about images.
\newblock In \emph{International Conference on Computer Vision (ICCV)}, 2015.

\bibitem[Mao et~al.(2015)Mao, Xu, Yang, Wang, Huang, and Yuille]{mao15iclr}
J.~Mao, W.~Xu, Y.~Yang, J.~Wang, Z.~Huang, and A.~Yuille.
\newblock Deep captioning with multimodal recurrent neural networks (m-rnn).
\newblock In \emph{International Conference on Learning Representations
  (ICLR)}, 2015.

\bibitem[Mao et~al.(2016)Mao, Huang, Toshev, Camburu, Yuille, and
  Murphy]{mao16cvpr}
J.~Mao, J.~Huang, A.~Toshev, O.~Camburu, A.~Yuille, and K.~Murphy.
\newblock Generation and comprehension of unambiguous object descriptions.
\newblock In \emph{Conference on Computer Vision and Pattern Recognition
  (CVPR)}, 2016.

\bibitem[Maron and Lozano-P{\'e}rez(1998)]{maron1998framework}
O.~Maron and T.~Lozano-P{\'e}rez.
\newblock A framework for multiple-instance learning.
\newblock \emph{Conference on Neural Information Processing Systems (NIPS)},
  1998.

\bibitem[Mensink et~al.(2012)Mensink, Verbeek, Perronnin, and
  Csurka]{mensink12eccv}
T.~Mensink, J.~Verbeek, F.~Perronnin, and G.~Csurka.
\newblock {Metric Learning for Large Scale Image Classification: Generalizing
  to New Classes at Near-Zero Cost}.
\newblock In \emph{European Conference on Computer Vision (ECCV)}, 2012.

\bibitem[Mihalcea and Moldovan(1999)]{mihalcea99cl}
R.~Mihalcea and D.~I. Moldovan.
\newblock A method for word sense disambiguation of unrestricted text.
\newblock In \emph{Proceedings of the Annual Meeting of the Association for
  Computational Linguistics (ACL)}, 1999.

\bibitem[Mikolov et~al.(2013)Mikolov, Sutskever, Chen, Corrado, and
  Dean]{mikolov13nips}
T.~Mikolov, I.~Sutskever, K.~Chen, G.~S. Corrado, and J.~Dean.
\newblock Distributed representations of words and phrases and their
  compositionality.
\newblock In \emph{Conference on Neural Information Processing Systems (NIPS)},
  2013.

\bibitem[Moses et~al.(1996)Moses, Ullman, and Edelman]{Moses1996}
Y.~Moses, S.~Ullman, and S.~Edelman.
\newblock {Generalization to novel images in upright and inverted faces}.
\newblock \emph{Perception}, 25:\penalty0 443--461, 1996.

\bibitem[Mrowca et~al.(2015)Mrowca, Rohrbach, Hoffman, Hu, Saenko, and
  Darrell]{mrowca15iccv}
D.~Mrowca, M.~Rohrbach, J.~Hoffman, R.~Hu, K.~Saenko, and T.~Darrell.
\newblock Spatial semantic regularisation for large scale object detection.
\newblock In \emph{International Conference on Computer Vision (ICCV)}, 2015.

\bibitem[Palatucci et~al.(2009)Palatucci, Pomerleau, Hinton, and
  Mitchell]{palatucci09nips}
M.~Palatucci, D.~Pomerleau, G.~Hinton, and T.~Mitchell.
\newblock Zero-shot learning with semantic output codes.
\newblock In \emph{Conference on Neural Information Processing Systems (NIPS)},
  2009.

\bibitem[Parikh and Grauman(2011)]{Parikh2011}
D.~Parikh and K.~Grauman.
\newblock {Relative attributes}.
\newblock In \emph{International Conference on Computer Vision (ICCV)}, 2011.

\bibitem[Plummer et~al.(2015)Plummer, Wang, Cervantes, Caicedo, Hockenmaier,
  and Lazebnik]{plummer15iccv}
B.~Plummer, L.~Wang, C.~Cervantes, J.~Caicedo, J.~Hockenmaier, and S.~Lazebnik.
\newblock Flickr30k entities: Collecting region-to-phrase correspondences for
  richer image-to-sentence models.
\newblock In \emph{International Conference on Computer Vision (ICCV)}, 2015.

\bibitem[Raina et~al.(2007)Raina, Battle, Lee, Packer, and Ng]{raina07icml}
R.~Raina, A.~Battle, H.~Lee, B.~Packer, and A.~Ng.
\newblock Self-taught learning: Transfer learning from unlabeled data.
\newblock In \emph{International Conference on Machine Learning (ICML)}, 2007.

\bibitem[Regneri et~al.(2013)Regneri, Rohrbach, Wetzel, Thater, Schiele, and
  Pinkal]{regneri13tacl}
M.~Regneri, M.~Rohrbach, D.~Wetzel, S.~Thater, B.~Schiele, and M.~Pinkal.
\newblock {Grounding Action Descriptions in Videos}.
\newblock \emph{Transactions of the Association for Computational Linguistics
  (TACL)}, 2013.

\bibitem[Ren et~al.(2015)Ren, Kiros, and Zemel]{ren15nips}
M.~Ren, R.~Kiros, and R.~Zemel.
\newblock Image question answering: A visual semantic embedding model and a new
  dataset.
\newblock In \emph{Conference on Neural Information Processing Systems (NIPS)},
  2015.

\bibitem[Rohrbach et~al.(2014)Rohrbach, Rohrbach, Qiu, Friedrich, Pinkal, and
  Schiele]{rohrbach14gcpr}
A.~Rohrbach, M.~Rohrbach, W.~Qiu, A.~Friedrich, M.~Pinkal, and B.~Schiele.
\newblock Coherent multi-sentence video description with variable level of
  detail.
\newblock In \emph{Proceedings of the German Confeence on Pattern Recognition
  (GCPR)}, 2014.

\bibitem[Rohrbach et~al.(2015{\natexlab{a}})Rohrbach, Rohrbach, Hu, Darrell,
  and Schiele]{rohrbach15arxiv1511.03745}
A.~Rohrbach, M.~Rohrbach, R.~Hu, T.~Darrell, and B.~Schiele.
\newblock Grounding of textual phrases in images by reconstruction.
\newblock \emph{arXiv preprint arXiv:1511.03745}, 2015{\natexlab{a}}.

\bibitem[Rohrbach et~al.(2015{\natexlab{b}})Rohrbach, Rohrbach, and
  Schiele]{rohrbach15gcpr}
A.~Rohrbach, M.~Rohrbach, and B.~Schiele.
\newblock The long-short story of movie description.
\newblock \emph{Proceedings of the German Confeence on Pattern Recognition
  (GCPR)}, 2015{\natexlab{b}}.

\bibitem[Rohrbach et~al.(2015{\natexlab{c}})Rohrbach, Rohrbach, Tandon, and
  Schiele]{rohrbach15cvpr}
A.~Rohrbach, M.~Rohrbach, N.~Tandon, and B.~Schiele.
\newblock A dataset for movie description.
\newblock In \emph{Conference on Computer Vision and Pattern Recognition
  (CVPR)}, 2015{\natexlab{c}}.

\bibitem[Rohrbach(2014)]{rohrbach14phd}
M.~Rohrbach.
\newblock \emph{Combining visual recognition and computational linguistics:
  linguistic knowledge for visual recognition and natural language descriptions
  of visual content}.
\newblock PhD thesis, Saarland University, 2014.

\bibitem[Rohrbach et~al.(2010)Rohrbach, Stark, Szarvas, Gurevych, and
  Schiele]{rohrbach10cvpr}
M.~Rohrbach, M.~Stark, G.~Szarvas, I.~Gurevych, and B.~Schiele.
\newblock {What helps Where - and Why? Semantic Relatedness for Knowledge
  Transfer}.
\newblock In \emph{Conference on Computer Vision and Pattern Recognition
  (CVPR)}, 2010.

\bibitem[Rohrbach et~al.(2011)Rohrbach, Stark, and Schiele]{rohrbach11cvpr}
M.~Rohrbach, M.~Stark, and B.~Schiele.
\newblock {Evaluating Knowledge Transfer and Zero-Shot Learning in a
  Large-Scale Setting}.
\newblock In \emph{Conference on Computer Vision and Pattern Recognition
  (CVPR)}, 2011.

\bibitem[Rohrbach et~al.(2012{\natexlab{a}})Rohrbach, Amin, Andriluka, and
  Schiele]{rohrbach12cvpr}
M.~Rohrbach, S.~Amin, M.~Andriluka, and B.~Schiele.
\newblock {A database for fine grained activity detection of cooking
  activities}.
\newblock In \emph{Conference on Computer Vision and Pattern Recognition
  (CVPR)}, 2012{\natexlab{a}}.

\bibitem[Rohrbach et~al.(2012{\natexlab{b}})Rohrbach, Regneri, Andriluka, Amin,
  Pinkal, and Schiele]{rohrbach12eccv}
M.~Rohrbach, M.~Regneri, M.~Andriluka, S.~Amin, M.~Pinkal, and B.~Schiele.
\newblock {Script data for attribute-based recognition of composite
  activities}.
\newblock In \emph{European Conference on Computer Vision (ECCV)},
  2012{\natexlab{b}}.

\bibitem[Rohrbach et~al.(2012{\natexlab{c}})Rohrbach, Stark, Szarvas, and
  Schiele]{rohrbach12lncs}
M.~Rohrbach, M.~Stark, G.~Szarvas, and B.~Schiele.
\newblock {Combining language sources and robust semantic relatedness for
  attribute-based knowledge transfer}.
\newblock In \emph{Proceedings of the European Conference on Computer Vision
  Workshops (ECCV Workshops)}, volume 6553 of \emph{{LNCS}},
  2012{\natexlab{c}}.

\bibitem[Rohrbach et~al.(2013{\natexlab{a}})Rohrbach, Ebert, and
  Schiele]{rohrbach13nips}
M.~Rohrbach, S.~Ebert, and B.~Schiele.
\newblock {Transfer Learning in a Transductive Setting}.
\newblock In \emph{Conference on Neural Information Processing Systems (NIPS)},
  2013{\natexlab{a}}.

\bibitem[Rohrbach et~al.(2013{\natexlab{b}})Rohrbach, Qiu, Titov, Thater,
  Pinkal, and Schiele]{rohrbach13iccv}
M.~Rohrbach, W.~Qiu, I.~Titov, S.~Thater, M.~Pinkal, and B.~Schiele.
\newblock Translating video content to natural language descriptions.
\newblock In \emph{International Conference on Computer Vision (ICCV)},
  2013{\natexlab{b}}.

\bibitem[Rohrbach et~al.(2015{\natexlab{d}})Rohrbach, Rohrbach, Regneri, Amin,
  Andriluka, Pinkal, and Schiele]{rohrbach15ijcv}
M.~Rohrbach, A.~Rohrbach, M.~Regneri, S.~Amin, M.~Andriluka, M.~Pinkal, and
  B.~Schiele.
\newblock Recognizing fine-grained and composite activities using hand-centric
  features and script data.
\newblock \emph{International Journal of Computer Vision (IJCV)},
  2015{\natexlab{d}}.

\bibitem[Senina et~al.(2014)Senina, Rohrbach, Qiu, Friedrich, Amin, Andriluka,
  Pinkal, and Schiele]{senina14arxiv}
A.~Senina, M.~Rohrbach, W.~Qiu, A.~Friedrich, S.~Amin, M.~Andriluka, M.~Pinkal,
  and B.~Schiele.
\newblock Coherent multi-sentence video description with variable level of
  detail.
\newblock \emph{arXiv preprint arXiv:1403.6173}, 2014.

\bibitem[Silberer et~al.(2013)Silberer, Ferrari, and Lapata]{silberer13acl}
C.~Silberer, V.~Ferrari, and M.~Lapata.
\newblock Models of semantic representation with visual attributes.
\newblock In \emph{Proceedings of the Annual Meeting of the Association for
  Computational Linguistics (ACL)}, 2013.

\bibitem[Simonyan and Zisserman(2015)]{simonyan15iclr}
K.~Simonyan and A.~Zisserman.
\newblock Very deep convolutional networks for large-scale image recognition.
\newblock In \emph{International Conference on Learning Representations
  (ICLR)}, 2015.

\bibitem[Sivic et~al.(2005)Sivic, Russell, Efros, Zisserman, and
  Freeman]{Sivic2005}
J.~Sivic, B.~C. Russell, A.~A. Efros, A.~Zisserman, and W.~T. Freeman.
\newblock {Discovering Object Categories in Image Collections}.
\newblock In \emph{International Conference on Computer Vision (ICCV)}, 2005.

\bibitem[Socher and Fei-Fei(2010)]{socher10cvpr}
R.~Socher and L.~Fei-Fei.
\newblock Connecting modalities: Semi-supervised segmentation and annotation of
  images using unaligned text corpora.
\newblock In \emph{Conference on Computer Vision and Pattern Recognition
  (CVPR)}, 2010.

\bibitem[S{\o}rensen(1948)]{sorensen48biol}
T.~S{\o}rensen.
\newblock A method of establishing groups of equal amplitude in plant sociology
  based on similarity of species and its application to analyses of the
  vegetation on danish commons.
\newblock \emph{Biol. Skr.}, 5:\penalty0 1--34, 1948.

\bibitem[Szegedy et~al.(2015)Szegedy, Liu, Jia, Sermanet, Reed, Anguelov,
  Erhan, Vanhoucke, and Rabinovich]{szegedy15cvpr}
C.~Szegedy, W.~Liu, Y.~Jia, P.~Sermanet, S.~Reed, D.~Anguelov, D.~Erhan,
  V.~Vanhoucke, and A.~Rabinovich.
\newblock Going deeper with convolutions.
\newblock In \emph{Conference on Computer Vision and Pattern Recognition
  (CVPR)}, 2015.

\bibitem[Thomee et~al.(2016)Thomee, Elizalde, Shamma, Ni, Friedland, Poland,
  Borth, and Li]{thomee2016yfcc100m}
B.~Thomee, B.~Elizalde, D.~A. Shamma, K.~Ni, G.~Friedland, D.~Poland, D.~Borth,
  and L.-J. Li.
\newblock Yfcc100m: The new data in multimedia research.
\newblock \emph{Communications of the ACM}, 59\penalty0 (2):\penalty0 64--73,
  2016.

\bibitem[Thrun(1996)]{thrun96nips}
S.~Thrun.
\newblock Is learning the n-th thing any easier than learning the first.
\newblock In \emph{Conference on Neural Information Processing Systems (NIPS)},
  1996.

\bibitem[Torabi et~al.(2015)Torabi, Pal, Larochelle, and
  Courville]{torabi15arxiv}
A.~Torabi, C.~Pal, H.~Larochelle, and A.~Courville.
\newblock Using descriptive video services to create a large data source for
  video annotation research.
\newblock \emph{arXiv preprint arXiv:1503.01070v1}, 2015.

\bibitem[Uijlings et~al.(2013)Uijlings, van~de Sande, Gevers, and
  Smeulders]{uijlings2013selective}
J.~R. Uijlings, K.~E. van~de Sande, T.~Gevers, and A.~W. Smeulders.
\newblock Selective search for object recognition.
\newblock \emph{International Journal of Computer Vision (IJCV)}, 104\penalty0
  (2):\penalty0 154--171, 2013.

\bibitem[Venugopalan et~al.(2015{\natexlab{a}})Venugopalan, Rohrbach, Donahue,
  Mooney, Darrell, and Saenko]{venugopalan15arxiv1505.00487v2}
S.~Venugopalan, M.~Rohrbach, J.~Donahue, R.~Mooney, T.~Darrell, and K.~Saenko.
\newblock Sequence to sequence -- video to text.
\newblock \emph{arXiv preprint arXiv:1505.00487v2}, 2015{\natexlab{a}}.

\bibitem[Venugopalan et~al.(2015{\natexlab{b}})Venugopalan, Rohrbach, Donahue,
  Mooney, Darrell, and Saenko]{venugopalan15iccv}
S.~Venugopalan, M.~Rohrbach, J.~Donahue, R.~Mooney, T.~Darrell, and K.~Saenko.
\newblock Sequence to sequence -- video to text.
\newblock In \emph{International Conference on Computer Vision (ICCV)},
  2015{\natexlab{b}}.

\bibitem[Venugopalan et~al.(2015{\natexlab{c}})Venugopalan, Xu, Donahue,
  Rohrbach, Mooney, and Saenko]{venugopalan15naacl}
S.~Venugopalan, H.~Xu, J.~Donahue, M.~Rohrbach, R.~Mooney, and K.~Saenko.
\newblock Translating videos to natural language using deep recurrent neural
  networks.
\newblock In \emph{Proceedings of the Conference of the North American Chapter
  of the Association for Computational Linguistics (NAACL)},
  2015{\natexlab{c}}.

\bibitem[Vinyals et~al.(2015)Vinyals, Toshev, Bengio, and Erhan]{vinyals15cvpr}
O.~Vinyals, A.~Toshev, S.~Bengio, and D.~Erhan.
\newblock Show and tell: {A} neural image caption generator.
\newblock In \emph{Conference on Computer Vision and Pattern Recognition
  (CVPR)}, 2015.

\bibitem[Wang and Schmid(2013)]{wang13iccv}
H.~Wang and C.~Schmid.
\newblock Action recognition with improved trajectories.
\newblock In \emph{International Conference on Computer Vision (ICCV)}, 2013.

\bibitem[Wang et~al.(2011)Wang, Kl{\"a}ser, Schmid, and Liu]{wang11cvpr}
H.~Wang, A.~Kl{\"a}ser, C.~Schmid, and C.-L. Liu.
\newblock {Action Recognition by Dense Trajectories}.
\newblock In \emph{Conference on Computer Vision and Pattern Recognition
  (CVPR)}, 2011.

\bibitem[Weber et~al.(2000)Weber, Welling, and Perona]{Weber2000}
M.~Weber, M.~Welling, and P.~Perona.
\newblock {Towards automatic discovery of object categories}.
\newblock In \emph{Conference on Computer Vision and Pattern Recognition
  (CVPR)}, 2000.

\bibitem[Wu et~al.(2016)Wu, Shen, Hengel, Wang, and Dick]{wu16arxiv1603.02814}
Q.~Wu, C.~Shen, A.~v.~d. Hengel, P.~Wang, and A.~Dick.
\newblock Image captioning and visual question answering based on attributes
  and their related external knowledge.
\newblock \emph{arXiv preprint arXiv:1603.02814}, 2016.

\bibitem[Yao et~al.(2015)Yao, Torabi, Cho, Ballas, Pal, Larochelle, and
  Courville]{yao15arxiv}
L.~Yao, A.~Torabi, K.~Cho, N.~Ballas, C.~Pal, H.~Larochelle, and A.~Courville.
\newblock Describing videos by exploiting temporal structure.
\newblock \emph{arXiv preprint arXiv:1502.08029v4}, 2015.

\bibitem[Young et~al.(2014)Young, Lai, Hodosh, and Hockenmaier]{young2014image}
P.~Young, A.~Lai, M.~Hodosh, and J.~Hockenmaier.
\newblock From image descriptions to visual denotations: New similarity metrics
  for semantic inference over event descriptions.
\newblock \emph{Transactions of the Association for Computational Linguistics
  (TACL)}, 2:\penalty0 67--78, 2014.

\bibitem[Zesch and Gurevych(2010)]{zesch09jnle}
T.~Zesch and I.~Gurevych.
\newblock Wisdom of crowds versus wisdom of linguists - measuring the semantic
  relatedness of words.
\newblock \emph{Natural Language Engineering}, 16\penalty0 (1):\penalty0
  25--59, 2010.

\bibitem[Zhou et~al.(2014)Zhou, Lapedriza, Xiao, Torralba, and
  Oliva]{zhou14nips}
B.~Zhou, A.~Lapedriza, J.~Xiao, A.~Torralba, and A.~Oliva.
\newblock {Learning Deep Features for Scene Recognition using Places Database.}
\newblock In \emph{Conference on Neural Information Processing Systems (NIPS)},
  2014.

\bibitem[Zhou et~al.(2004)Zhou, Bousquet, Lal, {Jason Weston}, and
  Sch\"{o}lkopf]{Zhou2004}
D.~Zhou, O.~Bousquet, T.~N. Lal, {Jason Weston}, and B.~Sch\"{o}lkopf.
\newblock {Learning with Local and Global Consistency}.
\newblock In \emph{Conference on Neural Information Processing Systems (NIPS)},
  2004.

\bibitem[Zhu et~al.(2003)Zhu, Ghahramani, and Lafferty]{Zhu2003}
X.~Zhu, Z.~Ghahramani, and J.~Lafferty.
\newblock {Semi-supervised learning using gaussian fields and harmonic
  functions}.
\newblock In \emph{International Conference on Machine Learning (ICML)}, 2003.

\bibitem[Zitnick et~al.(2013)Zitnick, Parikh, and Vanderwende]{zitnick13iccv}
C.~L. Zitnick, D.~Parikh, and L.~Vanderwende.
\newblock Learning the visual interpretation of sentences.
\newblock In \emph{International Conference on Computer Vision (ICCV)}, 2013.

\end{thebibliography}
